\documentclass[11pt]{article}

\usepackage[letterpaper,margin=1in,headheight=14pt]{geometry}
\usepackage[utf8]{inputenc}
\usepackage[T1]{fontenc}
\usepackage[sc]{mathpazo}
\usepackage{microtype}
\usepackage[dvipsnames]{xcolor}
\definecolor{accent}{RGB}{0,63,135}

\usepackage{amsmath,amssymb,amsthm}
\usepackage{booktabs,array,graphicx}
\usepackage{flafter}
\usepackage{placeins}
\usepackage[font=small,labelfont=bf,labelsep=period,skip=6pt]{caption}

\usepackage{titlesec}
\titleformat{\section}{\normalfont\Large\bfseries}{\color{accent}\thesection}{0.8em}{}
\titleformat{\subsection}{\normalfont\large\bfseries}{\color{accent}\thesubsection}{0.7em}{}
\titleformat{\subsubsection}{\normalfont\normalsize\bfseries}{\thesubsubsection}{0.6em}{}
\titleformat{\paragraph}[runin]{\normalfont\normalsize\bfseries}{}{0pt}{}
\titlespacing*{\section}{0pt}{2.4ex plus .6ex minus .2ex}{1.1ex plus .2ex}
\titlespacing*{\subsection}{0pt}{2.0ex plus .5ex minus .2ex}{0.9ex plus .2ex}
\titlespacing*{\subsubsection}{0pt}{1.6ex plus .4ex minus .2ex}{0.6ex plus .2ex}
\titlespacing*{\paragraph}{0pt}{1.3ex plus .4ex minus .2ex}{0.7em}

\usepackage{fancyhdr}
\renewcommand{\headrulewidth}{0.0pt}
\fancypagestyle{plain}{\fancyhf{}\fancyfoot[C]{\small\thepage}\renewcommand{\headrulewidth}{0pt}}

\usepackage[round]{natbib}

\usepackage{xurl}
\usepackage{hyperref}
\hypersetup{
  colorlinks=true,
  linkcolor=accent,
  citecolor=accent,
  urlcolor=accent,
  pdftitle={How Much Would a New Corpus Change the Prediction? Forecasting Corpus-Sampling Disagreement in Masked Text Reconstruction},
  pdfauthor={Hare Krishna}
}
\renewenvironment{abstract}{%
  \begin{list}{}{\leftmargin=0.07\textwidth\rightmargin=0.07\textwidth}%
  \item\relax\small
  {\centering\textsc{Abstract}\par\vspace{0.5em}}%
  \noindent\ignorespaces}{%
  \end{list}}

\newcommand{\E}{\mathbb E}
\newcommand{\Cov}{\operatorname{Cov}}

\newcommand{\tr}{\operatorname{tr}}

\newcommand{\cD}{\mathcal D}
\newcommand{\cA}{\mathcal A}

\newcommand{\Dbar}{\mathfrak D}
\theoremstyle{plain}
\newtheorem{proposition}{Proposition}

\begin{document}

\thispagestyle{plain}
\begin{center}
    {\LARGE\bfseries Forecasting Corpus-Sampling Disagreement\\[0.2em] in Masked Text Reconstruction\par}
  \vspace{1.6em}
  {\large Hare Krishna\par}
  \vspace{0.35em}
  {\itshape  University of Texas at Austin\par}
   {\small\texttt{hkrishna.phy@gmail.com}\par}
\end{center}
{\renewcommand{\thefootnote}{}\footnotetext{Code: \url{https://github.com/learncsai/masked-diffusion-model-text-reconstruction}}}
\vspace{0.6em}

\begin{abstract}
If a second text corpus were collected from the same population, how
much would a predictor's outputs change? We study this question for
masked-text reconstruction using a smoothed predictor built from nearby
token-pair counts and evaluated over the full 50,257-token GPT-2
vocabulary. Treating each source document as one observation, we derive
a first-order formula for the expected disagreement between predictors
fitted to two independent corpora. The formula accounts for variable
document length, random context counts, and correlations across token
offsets, and it can be evaluated from a separate reference corpus
without access to either target corpus. On TinyStories, WikiText-103,
and CNN/DailyMail, growing each corpus from 512 to 2,048 chunks of
64 tokens lowers disagreement by only 5--40\%, far short of the 75\%
implied by $1/n$ scaling. The forecast tracks this slow decline, which
we trace to smoothing of rarely seen contexts. Forecast accuracy depends
less on how these correlations are modeled than on the size of the
reference corpus: small references systematically underpredict
disagreement. A source-split jackknife that uses only reference data
reduces this calibration error by 57--59\%. After this correction, a
simpler smoothed-multinomial approximation is most accurate when the
reference corpus is smaller than the target corpus, and a nonlinear
source bootstrap is most accurate when it is larger. Finally, small
masked diffusion language models trained on the same corpora show that
the count-based forecasts partly identify which inputs are most
sensitive to the corpus, but miss the neural models' much faster decline
in disagreement.
\end{abstract}
\vspace{0.8em}

\section{Introduction and Related Work}
\label{sec:introduction}
A text corpus reflects the particular stories, articles, and other sources
that happened to be collected. A second collection from the same population
will contain different examples and token frequencies, so a predictor built
from it can change as well. How large should we expect that change to be?
Answering this question would help us decide how much text to collect and
judge how strongly a prediction depends on the chosen corpus. Average test
error alone cannot answer it: similarly accurate predictors may disagree
on the same input \citep{marx2020predictive}, and comparable validation
performance can conceal differences in deployment behavior
\citep{damour2022underspecification}.

We study how corpus sampling affects predictions in a masked-text
reconstruction problem. Our reconstructor predicts a hidden token by
combining conditional token frequencies at nearby positions. Independent
corpora A and B supply separate count tables, while the unigram baseline
and combination weights are fixed using shared auxiliary data. We measure
the observed disagreement as the mean squared distance between the two
reconstructors' output vectors on identical masked inputs. Our forecast
targets the expected value of this disagreement across independent draws
of A and B at a specified corpus size. We compute the forecast from an
independent reference corpus and assess it against the measured disagreement averaged over
multiple A/B pairs. Because only the count tables differ between the two
reconstructors, this comparison isolates variation due to corpus sampling.
Separately, we measure relative MSE improvement over the shared unigram
to assess reconstruction accuracy.

Two features of text make the calculation less straightforward than a
standard inverse-sample-size argument. First, token-pair counts are
dependent. An article can contribute several chunks, overlapping pairs
share tokens, and counts at different offsets fluctuate together.
Second, each conditional row is a ratio whose denominator, the number of
times the context appears, is itself random. Our smoothing rule pulls rows with
few observations strongly toward the unigram \citep{chen1999smoothing},
so a larger corpus does two things at once. It reduces sampling noise
in the estimated frequencies, and it lets the predictor rely more heavily
on those frequencies. The second effect partly offsets the first, which
is why disagreement can fall much more slowly than $1/n$.

\paragraph{Contributions.}
\textbf{(1)} We derive a first-order, source-level formula for expected
disagreement (Proposition~\ref{prop:law}) that accounts for source length,
random row totals, and cross-lag covariance. It enables forecasts from
independent reference data without using the target A/B counts or outputs,
and explains how smoothing can produce substantially slower decay than
$1/n$ at finite corpus sizes. We call it the source law throughout the article.

\textbf{(2)} We show that correcting finite-reference error can matter
more than retaining additional covariance detail. A source-split jackknife reduces small-reference error by 57--59\%
for both the source and smoothed multinomial forecasts.
In our main experiments, the corrected multinomial law outperforms the
source law with a small reference, while the corrected nonlinear source
bootstrap is most accurate with a larger one. In simulations with known population values, the multinomial law does better partly because two of its errors cancel.

\textbf{(3)} We test the count-based forecasts against disagreement between small
MDLMs trained on separate corpora. The forecasts partly identify the
inputs on which the MDLMs disagree most. We also compare the directions in which their token scores change. Overall token frequencies predict these directions as well as the reconstructor does. With the number of training steps held fixed, MDLM
disagreement falls much faster with corpus size than reconstructor
disagreement. The forecasts therefore help identify inputs sensitive
to corpus sampling, but do not fully explain the size or pattern of
neural prediction changes.

\paragraph{Related work.}
Existing work studies corpus uncertainty through embedding stability,
resampling, and covariance estimates for word representations
\citep{antoniak2018stability,gries2022uncertainty,vallebueno2024glovev}.
For word-frequency comparisons, \citet{lijffijt2016significance}
emphasize treating texts as independent observations while allowing
dependence among words within a text.
GloVe-V \citep{vallebueno2024glovev} instead defines uncertainty over
word co-occurrences rather than over the documents that produce them.
Our multinomial law makes the analogous count-level assumption, whereas
our source law treats complete sources as the sampling unit and
retains dependence within each source. Influence functions measure how single training examples change a model's predictions \citep{koh2017understanding} and \citet{brunet2019origins} trace embedding bias back to individual documents.
The infinitesimal jackknife linearizes a procedure with respect to
observation weights
\citep{jaeckel1972infinitesimal,efron1982jackknife,giordano2019swiss}.
\citet{wager2014confidence} use jackknife methods to estimate how much a random forest's prediction at a fixed input varies across training samples, a target close to ours.

Datamodels learn how a model's predictions depend on which examples it
is trained on, using outputs from models trained on many subsets of the
same dataset \citep{ilyas2022datamodels}. Our analytic forecast does not require training models on many subsets of the corpus: a first-order approximation gives an explicit variance formula for smoothed lag counts at a fixed chunk budget, and its inputs are estimated from an independent reference corpus. Treating a complete source as the sampling unit keeps
its internal dependence and lets us separate source-length,
row-normalization, and cross-lag contributions. We compare this formula
with a smoothed multinomial approximation and a nonlinear source bootstrap.
All three are computed from a finite reference corpus, so a forecast can
fail in two ways. Its approximation to the sampling variation can be too
crude, or the reference can estimate the required rows and covariances
poorly.

In our experiments, the second problem is the bigger one. Both analytic
laws (source law and multinomial) capture much of the slow, smoothing-driven decline in disagreement
with corpus size, but both underpredict its level when the reference is
small. Part of the gap comes from contexts the reference never sees,
which contribute nothing to the forecast even when the new corpora
contain them. We reduce this bias with a source-split jackknife
\citep{quenouille1956bias}, applied the same way to both laws. After
correction, the multinomial law is still the better calibrated of the two.
Exact population calculations explain this: even with no reference error, linearization makes the source formula overshoot the actual disagreement
by 3--12\% in our controlled experiment. The multinomial law gives smaller variance estimates that partly
offset this overshoot. As the reference grows and estimation error
shrinks, linearization error becomes a larger share of what remains.
This fits our finding that a corrected bootstrap, which refits the
smoothed rows on every resample instead of linearizing, gives the best
forecasts when the reference is larger than the target.

We then ask whether these local statistics also explain variation in a
trained neural denoiser. In image diffusion,
\citet{kadkhodaie2024generalization} show that denoisers trained on
non-overlapping datasets learn nearly the same score function when the
datasets are sufficiently large. \citet{zhang2024emergence} study
consistency of generated outputs across diffusion models under common
initial noise and deterministic sampling. \citet{wang2026rmt} connect
image-diffusion consistency to empirical means and covariances through
a Gaussian linear denoiser and random matrix theory. That Gaussian
analysis does not carry over directly to text, where tokens are
categorical, corruption is by masking, and the predictor divides by
random context counts. Table~\ref{tab:image-text-map} in
Appendix~\ref{app:image-text-comparison} maps out the correspondence.

For text, \citet{nguyen2024understanding} approximates transformer
predictions using training-data $n$-gram statistics and relates
approximation quality to prediction variability across training runs.
That variability comes from different shuffles of a fixed corpus;
our target is variation across independently sampled corpora.
Our matched experiment with small masked diffusion language models
(MDLMs) compares token-prediction vectors on identical masked inputs, and tests how far the count-based
predictions carry over in magnitude, input ranking, and output direction.

\section{Reconstructor}
\label{sec:task}

Our sampling unit is a source, such as a story, WikiText paragraph, or
news article, which may contribute several non-overlapping $L$-token
chunks. Each source belongs to exactly one data role
(Table~\ref{tab:data-roles}). Auxiliary sources set the shared unigram
and reconstruction weights, on which we condition throughout. Reference
sources then estimate sampling variation. The independent A and B
corpora supply the count tables whose disagreement we forecast, and
evaluation sources provide their common masked inputs. This separation
allows the forecast to be tested without using A/B outcomes to construct it.

\begin{table}[t]
\centering
\caption{Source-disjoint data roles. Auxiliary quantities are shared
by A and B and held fixed throughout.}
\label{tab:data-roles}
\small
\setlength{\tabcolsep}{6pt}
\renewcommand{\arraystretch}{1.08}
\begin{tabular}{@{}lp{0.55\textwidth}@{}}
\toprule
Role & Purpose \\
\midrule
Auxiliary & Set the unigram $p$ and reconstruction weights. \\
Reference & Estimate count variation and forecast disagreement. \\
A and B & Supply two independent count tables for testing the forecast. \\
Evaluation & Supply the same masked inputs to both predictors. \\
\bottomrule
\end{tabular}
\end{table}

Let $X=(X_0,\ldots,X_{L-1})$ have token IDs in $\{1,\ldots,K\}$.
Each position is hidden with probability $q$. The mask $M$ has hidden
set $H$, and $Z$ contains the visible tokens and mask. For output
vectors $u_i\in\mathbb R^K$, define
\[
 \|u\|_M^2=\frac{1}{\max\{1,|H|\}}\sum_{i\in H}\|u_i\|_2^2.
\]
This norm averages over hidden positions within a chunk, so every chunk
receives equal weight.

For one pair of corpora, define
\begin{equation}
 D_{A,B;q}=\E_{X,M}\|\widehat f_A(Z)-\widehat f_B(Z)\|_M^2.
\label{eq:observed}
\end{equation}
Averaging also over independent corpus collections gives the population
target $\Dbar_{n_A,n_B,q}=\E_{\cD_A,\cD_B}D_{A,B;q}$.
Here size counts number of chunks, and $\Dbar_{n,q}$ denotes equal
sizes $n_A=n_B=n$.

\paragraph{Reconstruction from lag counts.}
For offsets $0<|\delta|\le R$, let $S_{g,\delta}(b,a)$ count context
token $b$ followed by target $a$ at offset $\delta$ within the chunks
of source $g$. Both positions must be in the same chunk. Write
\begin{equation}
 N_{n,\delta}=\sum_g S_{g,\delta},\qquad
 \widehat C_n(\delta)=\frac{N_{n,\delta}}{n(L-|\delta|)}.
\label{eq:lags}
\end{equation}
These are raw pair frequencies.
With smoothing $\alpha>0$, the conditional row is
\begin{equation}
 \widehat P_\delta(a\mid b)=
 \frac{N_{n,\delta}(b,a)+\alpha p_a}{N_{n,\delta}(b,+)+\alpha}.
\label{eq:conditional}
\end{equation}
Here $N_{n,\delta}(b,+)=\sum_a N_{n,\delta}(b,a)$. Unseen contexts return $p$.
Let $\cA_i(M)$ contain the offsets with an in-bounds visible token at
$i-\delta$. The reconstruction method is
\begin{equation}
 \widehat f_i(Z)=p+\sum_{\delta\in\cA_i(M)}w_{\delta,\cA_i(M)}
 \{\widehat P_\delta(\cdot\mid X_{i-\delta})-p\}.
\label{eq:rule}
\end{equation}
Regularized least squares on auxiliary sources sets the weights once
(Appendix~\ref{app:computation}), leaving the count tables as the only
quantities that vary between A and B. With no visible neighbor, the
predictor returns $p$. For fixed masks and tables, its output is affine
in visible one-hot tokens. Its dependence on sampled counts is
nonlinear because each row is normalized. Because these scores can be negative, log-loss is undefined, so we use squared error.

\paragraph{Relative MSE improvement.}
For one-hot targets $e_{X_i}$, define
\begin{equation}
\begin{aligned}
 \mathcal L(f)&=\E_{X,M}\|f-e_X\|_M^2,\\
 \operatorname{RelImprove}(f)&=1-\frac{\mathcal L(f)}{\mathcal L(p)}.
\end{aligned}
\label{eq:rel-improve}
\end{equation}
The denominator is positive in every case we explore, so positive
relative improvement means lower MSE than the unigram. This metric uses
the full output vector. We assess ranking separately with top-$k$ accuracy
(Appendix~\ref{app:quality}). The unigram ignores context, so by construction it has zero disagreement and zero relative improvement.

\section{Disagreement Laws and Reference Correction}
\label{sec:theory}

A large corpus can contain many rare contexts, for which smoothing
strongly affects how predictions vary across corpus samples. In the
multinomial model below, the variance of a smoothed conditional row with
$N$ observations is proportional to $N/(N+\alpha)^2$. Increasing its
count from $N=\alpha$ to $N=4\alpha$ reduces this variance by only 36\%,
compared with 75\% under inverse-size scaling. So the forecast depends on how often each context is seen relative to $\alpha$. We first derive a source-level
formula that retains dependence among counts, then compare it with a
multinomial approximation to examine the roles of smoothing and covariance.

\subsection{Source influence and the population target}

Two independent corpora drawn by the same procedure at the same size
give identically distributed predictors. Expanding their squared
difference gives the exact identity
\begin{equation}
 \Dbar_{n,q}=2\E_{X,M}\E_{\cD}
 \|\widehat f_n(Z)-\E_{\cD}\widehat f_n(Z)\|_M^2.
\label{eq:identity}
\end{equation}
The remaining task is to approximate the variance of one predictor.
Appendix~\ref{app:proof} derives this identity and the first-order
covariance expansion.

Let $T_g$ be source $g$'s chunk contribution before terminal chunk-budget
truncation, and let $S_{g,\delta}$ count pairs in that full contribution.
Set $\tau=\E T_g$ and $s_\delta=\E S_{g,\delta}$. At chunk budget $n$, set
$G_n=n/\tau$ and $B_{\delta,n}=G_ns_\delta$. The centered source count is
\begin{equation}
 A_{g,\delta}=S_{g,\delta}-(T_g/\tau)s_\delta.
\label{eq:source}
\end{equation}
A longer source supplies more pairs but also uses more of the fixed
chunk budget, which is why centering subtracts $(T_g/\tau)s_\delta$.
When every source supplies one chunk, $T_g=\tau=1$ and this reduces
to ordinary mean centering.

Let $P_{\delta,n}$ be \eqref{eq:conditional} evaluated at counts
$B_{\delta,n}$. We expand around this fixed row, whose value generally
differs from $\E\widehat P_\delta$ because normalization is nonlinear.
Applying the quotient rule gives
\begin{equation}
 \dot P_{g,\delta,n}(a\mid b)=
 \frac{A_{g,\delta}(b,a)-P_{\delta,n}(a\mid b)A_{g,\delta}(b,+)}
 {B_{\delta,n}(b,+)+\alpha}.
\label{eq:row-influence}
\end{equation}
The subtraction keeps fluctuations in the row total. The resulting
output influence is
\begin{equation}
 h_{g,n}(i,Z)=\sum_{\delta\in\cA_i(M)}w_{\delta,\cA_i(M)}
 \dot P_{g,\delta,n}(\cdot\mid X_{i-\delta}).
\label{eq:output-influence}
\end{equation}
The vector \(h_{g,n}(i,Z)\) represents the first-order change in the reconstructed output at position \(i\) induced by source \(g\)'s count fluctuation, adjusted for the fixed chunk budget.

\begin{proposition}[First-order corpus reproducibility]
\label{prop:law}
Fix $K,L,R,q$, $\alpha>0$, and the shared unigram and auxiliary
weights. Suppose full source contributions are iid, with
$1\le T_g\le T_{\max}<\infty$, and $s_\delta(b,+)>0$ for every
row queried by the reconstructor.
For two independent corpora with $n$ retained chunks and
evaluation inputs independent of both,
\begin{equation}
 \Dbar_{n,q}=2G_n\E_{X,M,g}\|h_{g,n}(Z)\|_M^2+o(n^{-1})
\label{eq:law}
\end{equation}
as $n\to\infty$. The leading term is $O(n^{-1})$.
\end{proposition}

Appendix~\ref{app:proof} proves the result, including truncation
of the final source to meet the chunk budget. The finite-size
forecast retains the smoothed denominators in
\eqref{eq:row-influence}, so it can decrease more slowly than
$1/n$ while row counts remain comparable to $\alpha$.

\subsection{Forecast from independent reference data}

A reference has $G_{\rm ref}$ sources and $N_{\rm ref}$ chunks.
Replace the source means by their empirical values and use
$\widehat\tau=N_{\rm ref}/G_{\rm ref}$ and
$\widehat G_n=nG_{\rm ref}/N_{\rm ref}$. Then
\begin{equation}
 \widehat D^{\rm an}_{n,q}=
 \frac{2\widehat G_n}{G_{\rm ref}-1}
 \sum_{g=1}^{G_{\rm ref}}\E_{X,M}\|\widehat h_{g,n}(Z)\|_M^2.
\label{eq:reference}
\end{equation}
The forecast uses only reference and auxiliary quantities on the common
evaluation inputs.
The reference size determines how accurately we can estimate source moments and conditional rows, whereas the target size $n$ sets the expected counts, the strength of smoothing, and the number of sources in each new corpus. Forecasting another
$n$ therefore requires recomputing these quantities from the same
reference. The two sizes are free to differ, and
Section~\ref{sec:experiments} examines how their ratio affects calibration.

Source bootstrap estimates the same total variance by resampling whole
reference sources and rebuilding the nonlinear rows. The analytic
expression instead makes source and lag contributions explicit,
without bootstrap Monte Carlo error. We compare both forecasts in
Appendix~\ref{app:controls}.

\subsection{The smoothed multinomial approximation}

We compare the source forecast with a multinomial model that retains
smoothing but fixes row totals. For context $b$ at lag $\delta$, fix
$N=N_{n,\delta}(b)$ and let $C_a$ count targets of type $a$.
Assume $C=(C_1,\ldots,C_K)\sim\operatorname{Multinomial}(N,\pi)$,
where $\pi=\pi_\delta(\cdot\mid b)$ is the conditional target distribution.
The smoothed row $(C+\alpha p)/(N+\alpha)$ has covariance
$N\{\operatorname{diag}(\pi)-\pi\pi^\top\}/(N+\alpha)^2$.
For two independent corpora with the same row totals and independent
rows across lags, the predicted disagreement at a masked position is
\begin{equation}
\begin{aligned}
 v_{\delta,n}(b)&=\frac{2N_{n,\delta}(b)}{(N_{n,\delta}(b)+\alpha)^2}
 (1-\|\pi_\delta(\cdot\mid b)\|_2^2),\\
 d_n^{\rm multi}(i,Z)&=\sum_{\delta\in\cA_i(M)}
 w_{\delta,\cA_i(M)}^2 v_{\delta,n}(b_\delta),
\end{aligned}
\label{eq:multinomial-law}
\end{equation}
where $b_\delta=X_{i-\delta}$. Averaging over hidden positions and
chunks gives the total forecast. We estimate $\pi_\delta$ with
reference-smoothed rows and set $N_{n,\delta}(b)$ by scaling
reference row totals by $n/N_{\rm ref}$
(Appendix~\ref{app:matched-disagreement}).

The formula is exact under this model before substituting reference
estimates. The source law accounts for random row totals and dependence
from overlapping pairs, shared sources, and different lags, to first order.
Both include smoothing, so both can predict a decline slower than $1/n$. We compare how well each matches the level of disagreement and how it changes with corpus size.

\subsection{Correcting finite-reference bias}
\label{sec:correction}

The reference estimates both the conditional rows and their variation,
so increasing its size can change the forecast even when A/B size stays
fixed. We test a classical jackknife correction for this estimation
error \citep{quenouille1956bias}. Let $F_S(n)$ denote a forecast from
$S$ reference sources. Split them randomly into two groups of sizes
$S_1,S_2$, keeping each source's retained chunks together, and compute
\begin{equation}
 F_J(n)=2F_S(n)-\sum_{j=1}^2\frac{S_j}{S}F_{S_j}(n).
\label{eq:reference-jackknife}
\end{equation}
We use the same target size $n$ in all three components. The operation
applies to any forecast computed from a reference.
If the mean forecast has bias $b/S+o(S^{-1})$ at fixed source count,
\eqref{eq:reference-jackknife} cancels its leading term
(Appendix~\ref{app:jackknife}). This justifies the correction when the bias expansion is regular; whether it still helps with sparse rows and chunk-budget stopping is an empirical question, which we test below.

\section{Calibration and Corpus-Size Response}
\label{sec:experiments}
\label{sec:diagnosis}

We evaluate TinyStories \citep{eldan2023tinystories}, WikiText-103
\citep{merity2016pointer}, and cased CNN/DailyMail 3.0.0
\citep{hermann2015teaching,see2017get} with the GPT-2 tokenizer \citep{radford2019language}.
The shared settings are $K=50{,}257$, $L=64$, $R=2$, $\alpha=5$,
and auxiliary ridge $0.01$. The calibration comparisons below use $q=0.5$.
On the grid of three corpora, three sizes, three mask rates,
and two A/B pairs, the reconstructor improves on the unigram in
all 54 combinations, with mean relative MSE improvements of
13.16\%, 5.19\%, and 3.95\% in corpus order. Shuffling tokens within each chunk removes almost all of this gain (Appendix~\ref{app:quality}), so the gain comes from local context.

\subsection{\texorpdfstring{Disagreement falls more slowly than $1/n$}{Disagreement falls more slowly than 1/n}}

On seven independent pairs per corpus, increasing size from 512 to
2,048 chunks reduces disagreement by only 5 to 40\%, compared with
75\% under $1/n$. The source-law forecast, recomputed at each size, tracks this slow decline more closely than $1/n$ in every corpus. A separate three-pair
comparison at 4,096 and 8,192 chunks shows that the smoothed multinomial
law also reproduces much of the slow decrease
(Appendix~\ref{app:matched-size-response}). So smoothing, not covariance structure, is the main reason disagreement falls slowly. The absolute level is harder to predict, since it also depends on random row totals, source dependence, and
how well the reference estimates them.

\subsection{Bias correction improves calibration}

Our main calibration test uses 20 new A/B pairs and six separate
reference banks per corpus. Each A/B corpus contains
$n\in\{512,1024,2048,4096\}$ chunks, while each reference bank
contains $m\in\{1024,2048,4096\}$ chunks.
Smaller sizes are nested within each allocation. No source is shared between reference banks, between A/B corpora,
or across data roles. Appendix~\ref{app:correction-study}
gives the protocol and every tested condition. The primary comparison covers $(m,n)=(1024,2048),(1024,4096),(2048,4096)$,
where the reference is smaller than the target (A/B). For each method and reference bank, let $F$ denote the predicted
disagreement and $\bar D$ the arithmetic mean of the 20 observed
A/B disagreements. We measure the absolute log error
$|\log(F/\bar D)|$ and average equally over banks, $(m,n)$
conditions, and corpora.

\begin{table}[htbp]
\centering
\caption{Calibration on the twenty-pair, six-bank study. Lower mean absolute
log error is better. The primary subset contains the three $m<n$ cells.
The full grid contains all twelve $(m,n)$ combinations per corpus.
Jackknife rows apply correction at every cell, without the $m\le n$ restriction.
Appendix~\ref{app:added-baselines} gives definitions and uncertainty intervals.}
\label{tab:correction-primary}
\small
\setlength{\tabcolsep}{8pt}
\begin{tabular}{@{}lrr@{}}
\toprule
Forecast & $m<n\ \downarrow$ & Full grid $\downarrow$ \\
\midrule
Full source & 0.469 & 0.188 \\
Source jackknife & 0.203 & 0.103 \\
Multinomial & 0.205 & 0.115 \\
Multinomial jackknife & 0.085 & 0.061 \\
Source bootstrap & 0.433 & 0.191 \\
Bootstrap jackknife & 0.175 & 0.060 \\
\bottomrule
\end{tabular}
\end{table}

The source-split jackknife reduces the source forecast's error
by 56.7\%, from 0.469 to 0.203, bringing it close to the
uncorrected multinomial error of 0.205. Applying
the same correction to multinomial reduces its error by 58.5\% to
0.085 (Table~\ref{tab:correction-primary}).
The correction helps both laws without changing their order, and the simpler multinomial law remains the better calibrated of the two.
We also evaluate the correction on six fresh A/B pairs and three
new reference banks per corpus, saving forecasts before scoring.
At $m=1024,n=2048$, the same split jackknife reduces
multinomial error from 0.187 to 0.077
(Appendix~\ref{app:confirmation}).

\begin{figure}[t]
\centering
\includegraphics[width=\textwidth]{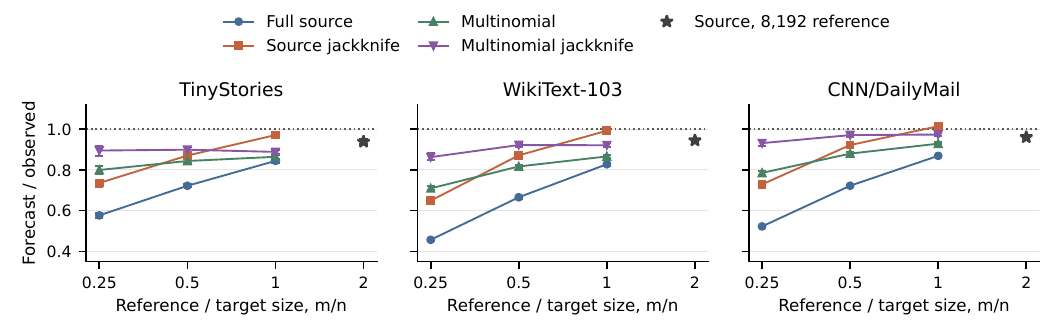}
\caption{Analytic forecast calibration at $n=4096$, with 64 tokens per chunk
and $q=0.5$. Each point divides the six-bank mean forecast by the
arithmetic mean disagreement of 20 A/B pairs. Bars show 95\% crossed
bootstrap intervals over banks and pairs, conditional on auxiliary and
evaluation data. The dotted line marks exact calibration, and lines
connect the three tested reference sizes. Stars use an uncorrected
8,192-chunk reference.}
\label{fig:reference-correction}
\end{figure}

Figure~\ref{fig:reference-correction} shows the effect of jackknife
correction on the analytic source forecast. At $m = n = 4096$, correction moves the ratios from 0.83--0.87 to 0.97--1.01. With $m = 1024$, the corrected ratios are still only 0.65--0.74. So correction reduces, but does not remove, the shortfall caused by a small reference.
For the analytic source law, we recommend correction only when $m\le n$:
correcting at $m>n$ raises source error from 0.052 to 0.095.

The nonlinear source bootstrap behaves differently.
At $m<n$, its corrected error is 0.175, compared with 0.085 for corrected
multinomial. At $m>n$, it improves from 0.072 to 0.013 after correction,
beating corrected multinomial's 0.049. On the separate confirmation pairs
at $m=2048,n=1024$, the corrected bootstrap again has lower error, 0.020
versus 0.049 (Appendix~\ref{app:bootstrap-confirmation}). We therefore favor corrected
multinomial with a small reference and corrected source bootstrap with a
larger one.

An obvious alternative is to scale an uncorrected forecast by a constant,
and it works well (Appendix~\ref{app:multiplier-transfer}). We also test whether a fitted scale factor transfers to fresh sources.
We choose a single multiplier to minimize the multinomial forecast's
mean absolute log error over the twenty-pair calibration grid.
Applying the resulting factor, $1.118$, unchanged to the fresh grid
gives error 0.045, compared with 0.060 for multinomial jackknife
(Appendix~\ref{app:multiplier-transfer}).
But fitting this factor needs observed A/B disagreements, while the jackknife needs only the reference bank.

\paragraph{Sensitivity to smoothing.}
The calibration ranking changes with the smoothing strength
(Table~\ref{tab:main-alpha-ablation}). Corrected multinomial has lower
error at $\alpha=1$ and 5, but at $\alpha=20$ corrected source has lower
error on TinyStories and WikiText, with the two methods nearly tied on
CNN/DailyMail.

\begin{table}[htbp]
\centering
\caption{Effect of $\alpha$ on calibration for small references ($m<n$).
Both forecasts use source-split jackknife correction. Entries are mean
absolute log errors ($\downarrow$) on the same 20 pairs and six reference
banks per corpus, with weights fixed at their $\alpha=5$ values.
Appendix~\ref{app:alpha-summary} gives the protocol and interpretation.}
\label{tab:main-alpha-ablation}
\small
\setlength{\tabcolsep}{6pt}
\begin{tabular}{@{}r*{6}{r}@{}}
\toprule
& \multicolumn{2}{c}{TinyStories} & \multicolumn{2}{c}{WikiText} & \multicolumn{2}{c}{CNN/DailyMail} \\
\cmidrule(lr){2-3}\cmidrule(lr){4-5}\cmidrule(l){6-7}
$\alpha$ & Source & Mult. & Source & Mult. & Source & Mult. \\
\midrule
1 & 0.525 & 0.220 & 0.803 & 0.307 & 0.677 & 0.260 \\
5 & 0.196 & 0.104 & 0.247 & 0.107 & 0.167 & 0.043 \\
20 & 0.052 & 0.108 & 0.057 & 0.114 & 0.024 & 0.024 \\
\bottomrule
\end{tabular}
\end{table}

\subsection{Where does the remaining error come from?}

To find where the remaining shortfall comes from, we use synthetic populations whose mean counts are
known exactly (Appendix~\ref{app:missing-contexts}). A context that never
appears in the reference contributes nothing to the forecast, however often
the A/B corpora contain it. With four chunks per source ($n=1024$, $m=128$),
these unseen rows and their cross terms account for 32.6 percentage points of
underprediction. Errors in rows the reference did observe add another 38.6,
so more than half of the shortfall lies in rows that a bias correction can
reach. When we replace every reference estimate with its exact population
value (Appendix~\ref{app:population-oracle}), the multinomial law is still
closer to the actual disagreement in all six $\alpha=5$ controls.
Linearization makes the source law overshoot, and the multinomial law's
smaller variance happens to cancel part of that. These are simulations, so real text may show different error sizes.

\section{Comparison with Trained MDLMs}
\label{sec:mdlm}
We next test whether the count-based forecasts say anything about neural models trained on the same corpora. We compare MDLM and reconstructor disagreement in overall size, in which inputs disagree most, and in the direction of the output differences.

We train MDLMs \citep{sahoo2024mdlm} with two Transformer blocks of width 96 on three A/B pairs at 512, 1,024, and 2,048 chunks. A and B share the initialization, training seed, masking streams, and number of updates, so only the training corpus differs. We score both at 8,000 and 12,000 updates on the common 128 chunks
with $q=0.5$. Appendix~\ref{app:mdlm-results} gives the full protocol
and seed controls.

Both models see the same corpora, masked inputs, and vocabulary, and we
score both with the squared-error measures of Section~\ref{sec:task},
applied to the full output vector. MDLM is trained with its usual likelihood loss. Two differences remain: It reads every visible position in the
64-token chunk and outputs probabilities, while the reconstructor reads only
the four nearest offsets and outputs affine scores that can be negative.

\paragraph{Magnitude of disagreements.}
From 512 to 2,048 chunks, MDLM disagreement falls by 72.9\%, 81.5\%,
and 79.9\% at 12,000 updates. The reconstructor falls by 38.9\%,
4.9\%, and 11.4\% on the same pairs. Thus the neural models have a much
steeper size response (Figure~\ref{fig:mdlm-disagreement}), and the
ordering of absolute disagreement across corpora also changes at
2,048 chunks (Table~\ref{tab:main-mdlm}).

\begin{table}[t]
\centering
\caption{Reconstruction at $n=2048$ chunks and mask rate $q=0.5$ for the
local reconstructor (L) and MDLM after 12,000 updates (N). For each corpus,
both methods are fitted to the A and B corpora of three independent pairs
and evaluated on the same 128 masked chunks. $D$ is the disagreement between
the A and B fits, averaged over the three pairs; lower means more stable.
RelImprove is the MSE improvement over the unigram, averaged over the six
fitted models; higher is better.}
\label{tab:main-mdlm}
\small
\setlength{\tabcolsep}{8pt}
\begin{tabular}{@{}lrrrr@{}}
\toprule
& \multicolumn{2}{c}{$D\downarrow$} & \multicolumn{2}{c}{RelImprove (\%) $\uparrow$} \\
\cmidrule(lr){2-3}\cmidrule(l){4-5}
Corpus & L & N & L & N \\
\midrule
TinyStories & 0.0223 & 0.0504 & 14.34 & 20.82 \\
WikiText-103 & 0.0308 & 0.0344 & 5.45 & 3.16 \\
CNN/DailyMail & 0.0248 & 0.0112 & 4.43 & 1.21 \\
\bottomrule
\end{tabular}
\end{table}

MDLM improves reconstruction more on TinyStories, but the reconstructor
performs better on WikiText and CNN/DailyMail. The news models are the most
stable neural models, yet they improve MSE over the unigram by only
1.21\%.
Training longer increased disagreement in all nine conditions.

\paragraph{Which inputs disagree most.}
Although the magnitudes differ, the count forecasts partly predict which
chunks the neural models disagree on most. Across the 128 evaluation chunks
at $n=2048$ and 12,000 updates, the Spearman correlation between the
source-law forecast and MDLM disagreement averages 0.48 on TinyStories, 0.38
on WikiText, and 0.18 on news. The multinomial law gives nearly identical values.

\paragraph{Direction of the output differences.}
\label{sec:mdlm-subspaces}
We next check whether the two models change in the same directions. We
project output differences onto 64 fixed random vocabulary directions. On
half of the evaluation sources, we find the four directions in which the
reconstructor's A/B differences are largest. On the other half, we measure
how much of MDLM's difference energy falls in them. The share is 39.4\% on
TinyStories, 56.0\% on WikiText, and 52.2\% on news, against 6.25\% for
random directions. But the leading directions of the unigram covariance
$\operatorname{diag}(p)-pp^\top$ capture slightly more (41.0\%, 58.9\%, and
58.2\%). So token frequencies alone explain the overlap
(Appendix~\ref{app:mdlm-subspaces})
and Appendix~\ref{app:quality} compares top-$k$ accuracy.

\section{Discussion and Limitations}
\label{sec:discussion}

Knowing how much a predictor depends on its training corpus helps judge how reproducible its predictions are and how much more data to collect. For a smoothed count-based
reconstructor, we show that a separate reference sample can forecast
disagreement between predictors fitted to new corpora. Our source-level
formula connects this variation to source length, random row totals,
and covariance across lags. Retaining smoothing also explains why
collecting more text can reduce disagreement more slowly than $1/n$
at the corpus sizes studied.

The experiments show that improving the reference estimates can matter
more than refining the covariance model. A source-split jackknife
reduces small-reference error by 57--59\% for both analytic laws,
using only reference data. Fitted multipliers are competitive, but
require earlier observed disagreements. In our tested settings,
the corrected multinomial law performs best when the reference is
smaller than the target, while the corrected source bootstrap performs
best when it is larger.

The source formula also explains why the simpler multinomial law can do better. In our simulations, the multinomial law leaves out some covariance terms, and its smaller forecast partly offsets the overshoot that linearization causes, though the balance changes with $\alpha$. The bootstrap avoids linearization altogether by refitting the smoothed rows on each resample.

\paragraph{Further applications.}
The approach could extend to other smoothed count statistics, including
$n$-gram models and normalized co-occurrence or pointwise mutual
information estimates \citep{chen1999smoothing,church1990word}.
Their derivatives would propagate source covariance into output
variation under suitable positivity and regularity conditions.
A further goal is to use these forecasts to guide corpus collection.
Evaluating candidate sizes could identify the smallest corpus expected
to meet a chosen stability target. Finer size grids and independently
chosen targets would allow us to test whether these forecasts can
support a concrete decision: how much text to collect for reproducible
predictions.

\paragraph{Limitations.}
Larger references need not improve calibration when reference and
linearization errors offset. Proposition~\ref{prop:law} assumes a fixed
vocabulary and positive context probabilities, while the jackknife
requires a regular bias expansion in the number of reference sources.
Neither argument gives uniform guarantees for sparse rows or growing
vocabularies. The correction also leaves unseen rows unresolved.
Accounting for their uncertainty requires assumptions about conditional
outputs and their joint variation, beyond missing mass alone.
Estimating that mass must itself account for sequence dependence
\citep{pananjady2024missing}.

Our intervals capture variation across reference banks and A/B pairs,
conditional on the auxiliary quantities and 128 evaluation chunks.
Six banks limit precision, and WikiText paragraph sampling leaves
dependence within articles unmodeled. The fresh-source check is small: six A/B pairs and three banks per corpus.

The MDLM comparison is only a first test of how local corpus statistics relate to neural variation. It covers small Transformers and
corpora, with three A/B pairs and fixed update budgets. Larger corpora
receive fewer passes,
so training also affects the measured size response.
Extending the comparison to larger models and datasets, with varied
training budgets, would help establish how much neural variation these
statistics can explain. Such tests should measure predictive quality
alongside stability and check whether directional predictions improve
on the token-frequency control.

\FloatBarrier

\section*{AI Use Statement}
Generative AI tools, including OpenAI Codex and Anthropic Claude,
were used during several stages of this work. These tools assisted
with preprocessing TinyStories, WikiText-103, and CNN/DailyMail
and constructing the text chunks described in the paper. They were also used in generating
code to implement the experiments, preparing tables and figures
from recorded experimental outputs. We have also used these tools in identifying relevant literature
and revising parts of the manuscript for clarity and readability.

The authors take full responsibility for the final manuscript
and supplementary materials, including the mathematical claims,
experimental results, code, and references.


\clearpage
\appendix
\begin{center}
  {\LARGE\bfseries Appendix\par}
\end{center}
\vspace{0.5em}

\section{Proof and Exact Identities}
\label{app:proof}

\paragraph{Disagreement identity.}
First we fix the auxiliary quantities and an evaluation input $Z$.
Let $Y_A$ and $Y_B$ be the vectors of predictions at the hidden
positions, each divided by $\sqrt{\max\{1,|H|\}}$. Then
$\|Y_A-Y_B\|^2$ is the squared output difference in
\eqref{eq:observed}. Under this conditioning, the two predictions are
independent. Writing their means as $\mu_A$ and $\mu_B$ and
expanding the square gives
\[
 \E\|Y_A-Y_B\|^2
 =
 \E\|Y_A-\mu_A\|^2
 +
 \E\|Y_B-\mu_B\|^2
 +
 \|\mu_A-\mu_B\|^2.
\]
At equal corpus sizes, the two outputs have the same distribution,
so their means agree and the two variance terms are equal.
This holds even if the reconstructor is biased.
The last term therefore vanishes, and averaging over evaluation
inputs gives \eqref{eq:identity}. When the corpus sizes differ,
the means need not agree; averaging the full expression gives
\eqref{eq:unequal}.

\paragraph{From count covariance to output covariance.}
We condition on the auxiliary quantities. Let $\widehat\theta_n$
be the vector containing all lag counts, and
$\Omega_n=\Cov_{\cD}(\widehat\theta_n)$.
For a fixed evaluation input $Z$, let $F_Z(\theta)$ contain the
predictions at hidden positions, divided by
$\sqrt{\max\{1,|H|\}}$, when the count vector is $\theta$.
Define
\[
 \theta_n^\circ=\E_{\cD}\widehat\theta_n,
 \qquad
 J_n(Z)=D_\theta F_Z(\theta_n^\circ).
\]
The first-order expression is
\begin{equation}
 \Dbar_{n,q}
 \simeq
 2\E_{X,M}\tr\{J_n(Z)\Omega_nJ_n(Z)^\top\}.
\label{eq:geometry}
\end{equation}
This follows from the multivariate delta method
\citep[Chapter~3]{vandervaart1998asymptotic}. Expanding around the mean count
vector gives
\[
 F_Z(\widehat\theta_n)-F_Z(\theta_n^\circ)
 \simeq
 J_n(Z)(\widehat\theta_n-\theta_n^\circ).
\]
The linear term has covariance
$J_n(Z)\Omega_nJ_n(Z)^\top$.
Its expected squared norm is the trace of this covariance, so
\eqref{eq:identity} gives \eqref{eq:geometry}.
We now derive the count covariance under source sampling and
show that the omitted terms are $o(n^{-1})$.

\paragraph{Source counts at a fixed chunk budget.}
Longer sources use more of the chunk budget, leaving room for
fewer subsequent sources. The resulting correction has the
residual form familiar from ratio estimation
\citep[Chapter~6]{cochran1977sampling}.

Let $(T_g,S_g)$ be iid contributions from complete sources,
where $T_g$ is the number of chunks and $S_g$ contains the
counts at all offsets. Assume $1\le T_g\le T_{\max}$
\[
 \tau=\E T_g,\qquad
 s=\E S_g,\qquad
 G_n=\frac{n}{\tau},\qquad
 B_n=G_ns.
\]
Sampling stops at
\[
 N(n)=\min\left\{m:\sum_{g=1}^{m}T_g\ge n\right\}.
\]
We shorten the final source if necessary to obtain exactly $n$
chunks. Since $L,R,T_{\max}$ are fixed, this changes the count
vector by a uniformly bounded amount. With
\[
 A_g=S_g-\frac{T_g}{\tau}s,
\]
we have
\[
 \widehat\theta_n-B_n
 =
 \sum_{g=1}^{N(n)}A_g+d_n,
 \qquad
 \|d_n\|\le C.
\]
The residual satisfies $\E A_g=0$.
$A_g/\tau$ is the influence
function of the count rate $s/\tau$
\citep{hampel1974influence}.

The event $\{N(n)\ge g\}$ depends only on sources preceding $g$.
Then, the stopped sum of the $A_g$ has mean zero and
covariance
\[
 \Cov\left(\sum_{g=1}^{N(n)}A_g\right)
 =
 \E N(n)\,\E_g[A_gA_g^\top].
\]
The bounded overshoot and Wald's identity \citep{wald1944cumulative} give
$\E N(n)=G_n+O(1)$.
Since $d_n$ is bounded, its covariance with the stopped sum
is $O(\sqrt n)$ by Cauchy-Schwarz. Hence
\[
 \Omega_n
 =
 G_n\E_g[A_gA_g^\top]+o(n).
\]
Using one source vector for all offsets includes correlations
between lags as well as dependence within each source.

Taking expectations also gives $\theta_n^\circ=B_n+O(1)$.
Under the positive-row assumption below, moving the expansion
point from $\theta_n^\circ$ to $B_n$ changes the derivative by
$O(n^{-2})$. Thus the derivatives evaluated at
$B_{\delta,n}=G_ns_\delta$ in \eqref{eq:row-influence}
give the same first-order result.

\paragraph{Ratio expansion.}
A perturbation $\epsilon A_{g,\delta}$ adds
$\epsilon A_{g,\delta}(b,a)$ to the smoothed numerator and
$\epsilon A_{g,\delta}(b,+)$ to its denominator.
Differentiating the ratio at $\epsilon=0$ gives
\eqref{eq:row-influence}.

We keep $\alpha>0$, $K,L,R,T_{\max}$, and the auxiliary
quantities fixed as $n$ grows. Assume that every queried
population row has positive mass. Its denominator at
$B_{\delta,n}$ is then proportional to $n$.
When the empirical denominator is also bounded below by a
positive constant times $n$, the first derivative is
$O(n^{-1})$ and the Hessian is $O(n^{-2})$ throughout the
line segment between the population and empirical counts.

To control small empirical denominators, fix a queried row
$(\delta,b)$ and write
\[
 Y_g=S_{g,\delta}(b,+),\qquad
 0\le Y_g\le C_\delta=T_{\max}(L-|\delta|),
 \qquad
 \mu_{\delta b}=\E Y_g>0.
\]
At least
\[
 m_n=\left\lceil\frac{n}{T_{\max}}\right\rceil-1
\]
sources are included in full before the final source.
The empirical row total is therefore at least
$\sum_{g=1}^{m_n}Y_g$.
These first $m_n$ variables are iid; their inclusion does not
depend on the later stopping index. Hoeffding's inequality \citep{hoeffding1963probability} gives
\[
 \Pr\left\{
   \sum_{g=1}^{m_n}Y_g<\frac12m_n\mu_{\delta b}
 \right\}
 \le
 \exp\left(
   -\frac{m_n\mu_{\delta b}^2}{2C_\delta^2}
 \right).
\]
For sufficiently large $n$, the empirical denominator is thus
bounded below by a positive constant times $n$, except on an
event with exponentially small probability. There are finitely
many queried rows because $K,L,R$ are fixed, so a union bound
gives the same conclusion for all of them.

The stopped count sum has bounded martingale increments and
at most $n$ terms. Its fourth moment is therefore $O(n^2)$,
and the bounded correction $d_n$ gives
\[
 \E_{\cD}\|\widehat\theta_n-B_n\|^4=O(n^2).
\]
On the event where all queried denominators are sufficiently
large, the Taylor remainder is bounded by
$Cn^{-2}\|\widehat\theta_n-B_n\|^2$.
On the exceptional event, both the predictions and their
linear approximation are bounded: smoothing bounds each
conditional row, the weights are fixed, and the count
perturbation is $O(n)$.
Together, these facts give an $O(n^{-2})$ bound on the
expected squared norm of the Taylor remainder.

The boundary correction contributes
$D_\theta F_Z(B_n)d_n$, whose norm is $O(n^{-1})$.
Including this term in the remainder, we obtain
\[
 \begin{aligned}
 F_Z(\widehat\theta_n)
 &=
 F_Z(B_n)
 +D_\theta F_Z(B_n)\sum_{g=1}^{N(n)}A_g
 +r_n(Z),\\
 \E_{\cD}\|r_n(Z)\|^2&=O(n^{-2}).
 \end{aligned}
\]
The bound is uniform over evaluation inputs under the stated
assumptions. Its constants depend on the positive population
row masses and need not remain bounded as those masses
approach zero.

\paragraph{Output variance.}
Applying the row derivative to the reconstruction rule
\eqref{eq:rule} gives $h_{g,n}$.
Let $u_{g,n}(Z)$ be its vector of hidden outputs with the same
normalization as $F_Z$. Then
\[
 u_{g,n}(Z)=D_\theta F_Z(B_n)A_g,
 \qquad
 \|u_{g,n}(Z)\|^2=\|h_{g,n}(Z)\|_M^2.
\]
For fixed $Z$, this contribution has mean zero and norm
$O(n^{-1})$. The same stopped-sum calculation gives
\[
 \begin{aligned}
 \Cov_{\cD}\left(
   \sum_{g=1}^{N(n)}u_{g,n}(Z)
 \right)
 &=
 \E N(n)\,\E_g[u_{g,n}(Z)u_{g,n}(Z)^\top]\\
 &=
 G_n\E_g[u_{g,n}(Z)u_{g,n}(Z)^\top]
 +O(n^{-2}).
 \end{aligned}
\]
The linear fluctuation has expected squared norm $O(n^{-1})$,
whereas $\E_{\cD}\|r_n(Z)\|^2=O(n^{-2})$.
Their mixed covariance is therefore $O(n^{-3/2})$ by
Cauchy-Schwarz. Taking the trace, averaging over evaluation
inputs, and using \eqref{eq:identity} gives
\[
 \Dbar_{n,q}
 =
 2G_n\E_{X,M}\E_g\|h_{g,n}(Z)\|_M^2
 +o(n^{-1}),
\]
which proves Proposition~\ref{prop:law}.

For unequal corpus sizes, the exact identity also contains
the difference between the two mean predictions:
\begin{equation}
 \Dbar_{n_A,n_B,q}
 =
 V_{n_A,q}+V_{n_B,q}+Q_{n_A,n_B,q},
\label{eq:unequal}
\end{equation}
where
\[
 V_{n,q}=\frac{\Dbar_{n,q}}{2},
 \qquad
 \mu_n(Z)=\E_{\cD}\widehat f_n(Z),
\]
and
\[
 Q_{n_A,n_B,q}
 =
 \E_{X,M}\|\mu_{n_A}(Z)-\mu_{n_B}(Z)\|_M^2.
\]
Our first-order prediction adds the two variance estimates:
\[
 \widehat D^{\rm an}_{n_A,n_B,q}
 =
 \frac{
   \widehat D^{\rm an}_{n_A,q}
   +\widehat D^{\rm an}_{n_B,q}
 }{2}.
\]
The mean term is lower order under the assumptions above,
as we show next. It is still part of the finite-size target.
For equal sizes, $Q_{n,n,q}=0$ exactly.

\paragraph{Unequal-size mean order.}
Let $f_\infty$ be the reconstruction obtained from the
population conditional rows without smoothing.
Because $\alpha$ is fixed and the queried row masses are
positive, the prediction evaluated at $B_n$ differs from
$f_\infty$ by $O(n^{-1})$.
The stopped linear term has mean zero, and
\[
 \|\E_{\cD}r_n(Z)\|
 \le
 \bigl(\E_{\cD}\|r_n(Z)\|^2\bigr)^{1/2}
 =
 O(n^{-1}).
\]
It follows that
\[
 \|\mu_n(Z)-f_\infty(Z)\|_M=O(n^{-1})
\]
uniformly over evaluation inputs. Hence
\[
 \begin{aligned}
 Q_{n_A,n_B,q}
 &=
 O\bigl((n_A^{-1}+n_B^{-1})^2\bigr)\\
 &=
 o(n_A^{-1}+n_B^{-1})
 \end{aligned}
\]
as both sizes grow. Thus the mean term is smaller than the
$n_A^{-1}+n_B^{-1}$ variance scale.

\paragraph{Multinomial variance.}
Under the fixed-count model in Section~\ref{sec:theory},
a row count vector $C$ is the sum of $N$ independent one-hot
draws with mean $\pi$. Therefore
\[
 \Cov(C)
 =
 N\{\operatorname{diag}(\pi)-\pi\pi^\top\}
\]
\citep[Chapter~1]{agresti2013categorical}.
Adding the fixed vector $\alpha p$ does not change the
covariance, while division by $N+\alpha$ multiplies it by
$(N+\alpha)^{-2}$.
The difference between two independent copies has twice this
covariance, with trace
\[
 \frac{2N}{(N+\alpha)^2}
 \bigl(1-\|\pi\|^2\bigr).
\]
This gives the row factor in \eqref{eq:multinomial-law}.
Under the assumed independence across lags, the cross terms
in the squared output difference have zero expectation.
The formula is therefore exact for this model when evaluated
at its population quantities. Substituting reference estimates
introduces a separate estimation error.

\paragraph{Source-split jackknife.}
\label{app:jackknife}
Fix the target size $n$ and condition on the auxiliary
quantities and evaluation inputs.
Suppose the forecast from $S$ iid sources satisfies
\[
 \E F_S=F_\infty+\frac{b}{S}+r_S,
 \qquad
 r_S=o(S^{-1}).
\]
Partition the sources independently of their contents into
groups of deterministic sizes $S_1,S_2$, where
\[
 S_1+S_2=S,
 \qquad
 \frac{S_j}{S}\longrightarrow c_j\in(0,1).
\]
Let $F_{S_j}^{(j)}$ be the forecast computed from group $j$,
and define
\[
 F_J
 =
 2F_S-\sum_{j=1}^2\frac{S_j}{S}F_{S_j}^{(j)}.
\]
Each group has the marginal distribution of an iid sample
of its stated size. Linearity of expectation therefore gives
\[
 \begin{aligned}
 \E F_J
 &=
 F_\infty+\frac{2b}{S}
 -\sum_{j=1}^2\frac{S_j}{S}\frac{b}{S_j}
 +2r_S-\sum_{j=1}^2\frac{S_j}{S}r_{S_j}\\
 &=
 F_\infty+o(S^{-1}).
 \end{aligned}
\]
Both group estimates share data with the full-bank estimate,
but this expectation calculation requires no independence
between them. The weights $S_j/S$ cancel the leading bias
even when the groups differ in size. Splitting individual
chunks would change the sampling unit and would no longer
match this iid-source argument.

This is the classical bias-reduction argument of
\citet{quenouille1956bias}, applied to the reference forecast.
Its use for ratio estimators has a direct precedent in
\citet{durbin1959ratios}.
Here $F_\infty$ is the limiting reference forecast at the
chosen target size $n$. It can still differ from the exact
disagreement of the nonlinear reconstructor.

The calculation assumes a fixed number of iid sources and
a regular bias expansion. Our reference banks end at a fixed
chunk budget, so their source counts are random and their
final sources may be shortened. The calculation above does
not by itself establish the same bias order for that sampling
rule; we evaluate the correction under that protocol in the
experiments.

A row absent from the whole bank is also absent from both
groups, so the correction cannot recover its contribution
to the influence variance. We keep the signs of individual
contributions and set only a negative final scalar forecast
to zero. The bias expansion above applies before this final
clipping.

\FloatBarrier
\section{Data Protocol and Calibration Evidence}
\label{app:calibration-details}

\subsection{Finite-reference correction}
\label{app:correction-study}

For each corpus, we use a block of 128,000 distinct texts after the
64,000-source prefix used in the previous experiments (see Appendix~\ref{app:earlier-studies}). From this block,
we construct 20 source-disjoint A/B pairs and six independent reference
banks. Each corpus in a pair contains 4,096 chunks, and each reference
bank contains 8,192 chunks. Smaller sizes are obtained from prefixes.
Source IDs, text hashes, and chunk hashes verify separation between
A and B and across data roles. The auxiliary statistics and 128
evaluation chunks are shared with the matched comparison.

We write $m$ for the reference size and $n$ for the size of each corpus
in an A/B pair. Throughout this experiment, $q=0.5$, $R=2$, and
$\alpha=5$, with a ridge penalty of $0.01$. At each reference size,
a random permutation partitions the sources into two groups,
with all included chunks from a source assigned to the same group.
We recompute the conditional rows and covariance estimates in each
group for the same target size $n$, then combine the forecasts using
the source-count weights in \eqref{eq:reference-jackknife}.
The correction is evaluated at $m\in\{1024,2048,4096\}$.
The uncorrected forecast at $m=8192$ provides a comparison with a
larger reference bank.

The correction was checked in three simulation controls at $\alpha=5$
before the text experiment. The full source law, its jackknife
correction, and the multinomial law were specified before the A/B
outcomes were inspected. The multinomial correction and the additional
baselines were examined after the outcomes were available.
Section~\ref{app:confirmation} evaluates the multinomial correction
on a disjoint source allocation, with forecasts saved before scoring.

\paragraph{Error measure and uncertainty.}
Let $F_{c,b,m,n}^{h}$ denote the forecast from method $h$ for corpus
$c$, reference bank $b$, reference size $m$, and target size $n$.
Let $\bar D_{c,n}$ be the mean observed disagreement across the
20 A/B pairs. For reference banks smaller than the target corpora,
we summarize error over
\[
 \mathcal P=\{(1024,2048),(1024,4096),(2048,4096)\}
\]
as
\[
 E^h=
 \frac{1}{3\cdot3\cdot6}
 \sum_c\sum_{(m,n)\in\mathcal P}\sum_{b=1}^{6}
 \left|\log\frac{F_{c,b,m,n}^{h}}{\bar D_{c,n}}\right|.
\]
The three corpora, three size combinations, and six reference banks
receive equal weight.

We estimate uncertainty with 5,000 bootstrap draws. Within each
corpus, a draw resamples six bank indices and twenty pair indices
independently. The same indices are then used for every method and
every reference and target size. We recompute the mean observed
disagreements, the errors, and their paired differences in each draw.
This accounts for the dependence between methods and between
measurements at nested sizes. The intervals reflect variation across reference banks and A/B pairs only; the corpora, auxiliary data, evaluation inputs, and masks are held fixed.

\begin{figure}[htbp]
\centering
\includegraphics[width=0.94\textwidth]{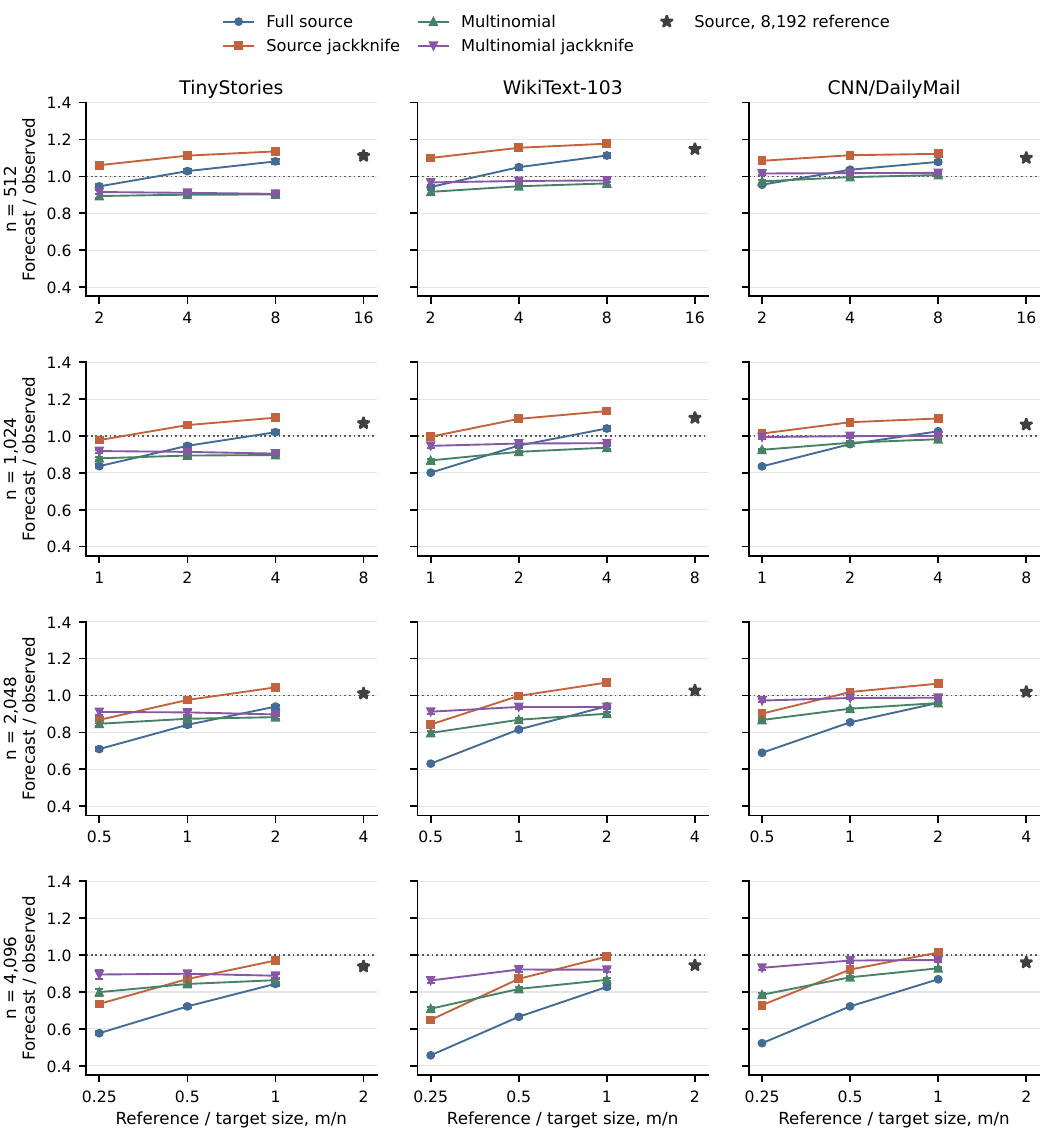}
\caption{Forecast calibration across reference and target sizes.
Rows correspond to target size $n$ and columns to corpora; sizes
count 64-token chunks. Each point is the mean forecast across six
reference banks divided by the mean observed disagreement across
20 A/B pairs. The dotted line marks exact calibration, with values
below one indicating underprediction and values above one indicating
overprediction. Error bars are 95\% intervals obtained by resampling
banks and A/B pairs independently, with auxiliary data, evaluation
inputs, and masks fixed. Curves use reference sizes
$m\in\{1024,2048,4096\}$; stars show the uncorrected forecast at
$m=8192$.}
\label{fig:correction-all}
\end{figure}

Table~\ref{tab:correction-primary} reports the errors over
$\mathcal P$. The paired difference between source jackknife and
the uncorrected source law is
$-0.265\ [-0.267,-0.263]$.
The difference between source jackknife and uncorrected multinomial
is $-0.0017\ [-0.0052,0.0020]$.
Applying the correction to the multinomial law reduces error further:
source-jackknife error minus multinomial-jackknife error is
$0.119\ [0.114,0.124]$.
Figure~\ref{fig:correction-all} shows the full grid, including $m>n$, where some corrections overpredict.

\subsection{Bootstrap and calibration baselines}
\label{app:added-baselines}
\label{app:controls}
We compare additional forecasts on the same twenty-pair data.
These comparisons were designed after the A/B outcomes were available.
Before computing these forecasts, we fixed the number of bootstrap
resamples, the division of reference sources into two groups, and the
multiplier $1.25$ used to test a uniform 25\% increase in the uncorrected
source and multinomial forecasts.

\paragraph{Source bootstrap and its correction.}
Following the cluster bootstrap
\citep{efron1979bootstrap,field2007clustered},
we sample sources uniformly with replacement until reaching $n$
chunks, shortening only the final source when necessary.
Each sampled source contributes its chunks and all associated
pair counts at every lag. We rebuild the smoothed conditional
rows on each draw and estimate disagreement as twice the unbiased
sample output variance on the common evaluation inputs.

Three independent batches of sixteen replicates give 48 replicates
per forecast. To apply the jackknife, we repeat the procedure on
the two source groups used by the analytic corrections and combine
the resulting scalar estimates according to
\eqref{eq:reference-jackknife}.
Across 216 forecasts per method, the median Monte Carlo standard
errors are 0.62\% for the bootstrap and 1.19\% for its jackknife
correction; the maxima are 2.36\% and 4.17\%.
All corrected scalar forecasts are positive.
The intervals below condition on these saved Monte Carlo draws.

\paragraph{Direct disagreement and multiplicative corrections.}
A second approach constructs a reconstructor from each source
group and measures their disagreement, $D_{\rm split}$.
If the groups contain $m_1$ and $m_2$ chunks, output variance
proportional to inverse corpus size and a size-independent mean
would imply
\[
 F_{\rm direct}(n)=D_{\rm split}\frac{h}{n},
 \qquad
 h=\frac{2}{1/m_1+1/m_2}.
\]
The harmonic mean $h$ accounts for unequal group sizes.
Since smoothing can change both the mean and the variance,
we also compare the constant forecast $D_{\rm split}$.

We then ask whether a simple increase in scale can account for
the improvement from the jackknife. For this comparison, we
multiply each uncorrected law by 1.25.
Finally, we consider a conditional correction: apply the jackknife
when $m\le n$ and use the uncorrected forecast when $m>n$.

Table~\ref{tab:followup-intervals} reports errors for the three
$m<n$ combinations, the full twelve-combination grid, and the six
$m>n$ combinations. Restricting the jackknife to $m\le n$ reduces
the full-grid error of the source law from 0.103 to 0.081.
The same restriction increases multinomial error from 0.061 to
0.071. So whether to restrict the correction depends on the law. A fixed multiplier also does well when $m < n$ but worse when $m > n$; Appendix~\ref{app:confirmation} tests whether fitted multipliers transfer.

\begin{table}[!htbp]
\centering
\caption{Mean absolute log prediction error across reference banks,
size combinations, and corpora, with 95\% bootstrap intervals.
Each forecast is compared with the mean disagreement of the same
twenty A/B pairs, and every method uses the same six reference banks
per corpus. Lower values indicate better calibration.
The $m<n$ column averages the three combinations in $\mathcal P$;
the full grid averages all twelve combinations; and the $m>n$
column averages the six combinations with a larger reference bank
than target corpus. Banks, combinations, and corpora receive equal
weight within each column. The last two rows apply the jackknife
only when $m\le n$ and use the uncorrected law otherwise.
Intervals resample whole banks and A/B pairs jointly across methods
and sizes, conditional on auxiliary data, evaluation inputs, masks,
and saved Monte Carlo draws.}
\label{tab:followup-intervals}
\small
\setlength{\tabcolsep}{6pt}
\begin{tabular}{@{}lrrr@{}}
\toprule
Forecast & $m<n$ & Full grid & $m>n$ \\
\midrule
Full source & 0.469 [0.463, 0.475] & 0.188 [0.185, 0.191] & 0.052 [0.051, 0.053] \\
Source jackknife & 0.203 [0.197, 0.211] & 0.103 [0.101, 0.105] & 0.095 [0.088, 0.100] \\
Multinomial & 0.205 [0.199, 0.212] & 0.115 [0.112, 0.119] & 0.069 [0.065, 0.073] \\
Multinomial jackknife & 0.085 [0.077, 0.093] & 0.061 [0.057, 0.065] & 0.049 [0.045, 0.053] \\
Source bootstrap & 0.433 [0.427, 0.440] & 0.191 [0.186, 0.195] & 0.072 [0.067, 0.076] \\
Bootstrap jackknife & 0.175 [0.167, 0.184] & 0.060 [0.056, 0.064] & 0.013 [0.011, 0.016] \\
Direct split, $1/n$ & 1.318 [1.311, 1.324] & 0.671 [0.669, 0.674] & 0.402 [0.398, 0.406] \\
Direct split, constant & 0.300 [0.293, 0.306] & 0.147 [0.144, 0.149] & 0.076 [0.073, 0.081] \\
Source $\times1.25$ & 0.245 [0.240, 0.251] & 0.184 [0.181, 0.186] & 0.222 [0.217, 0.227] \\
Multinomial $\times1.25$ & 0.054 [0.051, 0.057] & 0.117 [0.114, 0.120] & 0.155 [0.151, 0.159] \\
Source jackknife, $m\le n$ & 0.203 [0.197, 0.211] & 0.081 [0.079, 0.084] & 0.052 [0.051, 0.053] \\
Multinomial jackknife, $m\le n$ & 0.085 [0.077, 0.093] & 0.071 [0.067, 0.075] & 0.069 [0.065, 0.073] \\
\bottomrule
\end{tabular}
\end{table}

\subsection{Calibration on disjoint source collections}
\label{app:confirmation}

To assess the multinomial correction on different corpus pairs
and reference banks, we allocate unused texts from the same
128,000-text block to six A/B pairs and three reference banks
per corpus. Each A or B corpus and each bank contains 2,048
chunks, with a 1,024-chunk prefix.
Source IDs, text hashes, and chunk hashes verify separation
across roles and from the other source allocations.
The auxiliary quantities and evaluation inputs are unchanged.
This allocation uses 10,638 TinyStories sources, 13,556 WikiText
paragraphs, and 15,376 news articles.

Before allocating these sources, we specified the comparison
between multinomial and jackknife-corrected multinomial forecasts
at $m=1024,n=2048$. We use the same bank-level absolute log error
and 5,000 bootstrap draws as in the twenty-pair experiment. We also report the other three combinations of
$m,n\in\{1024,2048\}$ and the source-law forecasts.
The multiplier and bootstrap comparisons were designed after
these outcomes were available.

\begin{table}[!htbp]
\centering
\caption{Calibration on six A/B pairs and three reference banks
per corpus. Entries are mean absolute log prediction errors with
95\% bootstrap intervals; lower values indicate better calibration.
Sizes count 64-token chunks.
The $m<n$ column corresponds to $(m,n)=(1024,2048)$, the full
grid averages all four combinations of $m,n\in\{1024,2048\}$,
and the $m>n$ column corresponds to $(m,n)=(2048,1024)$.
Each forecast is compared with the mean disagreement across the
six pairs, with banks, size combinations, and corpora weighted
equally. Jackknife corrections are applied at every combination.
The pooled and cross-corpus multipliers are selected using only
the twenty-pair results. Intervals condition on auxiliary data, evaluation
inputs, masks, and, where applicable, saved Monte Carlo draws.}
\label{tab:confirmation}
\small
\setlength{\tabcolsep}{6pt}
\begin{tabular}{@{}lrrr@{}}
\toprule
Forecast & $m<n$ & Full grid & $m>n$ \\
\midrule
Full source & 0.397 [0.385, 0.410] & 0.207 [0.196, 0.217] & 0.053 [0.043, 0.063] \\
Source jackknife & 0.146 [0.133, 0.159] & 0.066 [0.060, 0.073] & 0.071 [0.059, 0.083] \\
Multinomial & 0.187 [0.175, 0.199] & 0.126 [0.118, 0.135] & 0.080 [0.073, 0.088] \\
Multinomial jackknife & 0.077 [0.066, 0.089] & 0.060 [0.051, 0.069] & 0.049 [0.041, 0.059] \\
Multinomial $\times1.25$ & 0.041 [0.031, 0.051] & 0.098 [0.090, 0.106] & 0.143 [0.135, 0.150] \\
Multinomial, pooled factor & 0.076 [0.062, 0.090] & 0.045 [0.039, 0.052] & 0.039 [0.031, 0.047] \\
Multinomial, cross-corpus factors & 0.082 [0.069, 0.095] & 0.057 [0.051, 0.064] & 0.052 [0.044, 0.061] \\
\bottomrule
\end{tabular}
\end{table}

At $m=1024,n=2048$, the multinomial correction reduces mean
absolute log error from 0.187 to 0.077.
The paired difference, corrected minus uncorrected, is
$-0.110\ [-0.113,-0.106]$, a 58.9\% reduction.
The corrected forecast-to-observation ratios are 0.897, 0.913,
and 0.970 for TinyStories, WikiText-103, and CNN/DailyMail,
respectively. So the improvement holds on new sources, although the corrected forecasts still run low. Both experiments draw from the same three corpora.

\paragraph{Bootstrap comparison at $m>n$.}
\label{app:bootstrap-confirmation}
We evaluate the source bootstrap at $m=2048,n=1024$ using all
three banks and six pairs per corpus.
The resampling procedure and source-split correction follow
Section~\ref{app:added-baselines}, with three batches of sixteen
replicates per component and the same source partitions as the
analytic corrections.

The correction reduces bootstrap error from 0.106 to 0.020.
Its paired error difference from multinomial jackknife is
$-0.030\ [-0.037,-0.021]$, and its difference from the multinomial
forecast with the pooled calibration factor is
$-0.019\ [-0.032,-0.006]$.
The corrected bootstrap forecast-to-observation ratios are
0.964, 0.987, and 1.001 for TinyStories, WikiText-103, and
CNN/DailyMail, respectively.
Its lower pooled error does not imply an advantage in every
corpus: the calibrated multiplier gives closer predictions
on TinyStories.

The median Monte Carlo standard errors are 0.61\% for the
bootstrap and 1.21\% for its correction, with maxima of 1.42\%
and 2.40\%. All nine corrected scalar estimates are positive.
The bank-and-pair intervals condition on these Monte Carlo
draws and on the shared auxiliary and evaluation data.

The results also suggest combining the two corrections.
On the twenty-pair grid, using multinomial jackknife when
$m\le n$ and bootstrap jackknife when $m>n$ gives error
$0.043\ [0.040,0.047]$.
This rule was chosen after inspecting those results.
The six-pair bootstrap comparison evaluates its $m>n$
component at $(m,n)=(2048,1024)$; it does not evaluate the
rule across the full grid or establish $m=n$ as a general
threshold.

\paragraph{Transfer of multiplicative corrections.}
\label{app:multiplier-transfer}
A fixed multiplier of 1.25 gives lower multinomial error than
the jackknife on the $m<n$ subset of the twenty-pair experiment.
The same comparison at $m=1024,n=2048$ in the six-pair experiment
gives errors of 0.041 and 0.077, with paired difference
$-0.036\ [-0.055,-0.017]$.
Across the four-combination grid, however, the multiplier gives
error 0.098, compared with 0.060 for the jackknife.
We therefore test whether a factor chosen on one set of results still works on new sources, corpora, and sizes.

\begin{figure}[htbp]
\centering
\includegraphics[width=0.96\textwidth]{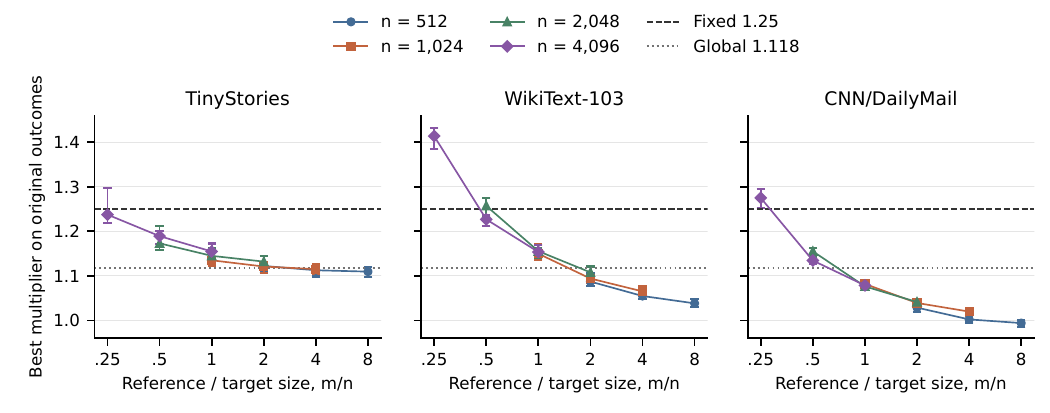}
\caption{Optimal multiplicative correction for the multinomial
forecast at each reference--target size combination in the
twenty-pair experiment. Each factor minimizes mean absolute log
error across six reference banks, using the mean observed
disagreement of twenty A/B pairs.
Shapes identify target size $n$, and lines connect combinations
with the same target size.
Error bars are 95\% intervals obtained by resampling banks and
A/B pairs and selecting the factor again in each draw.
The dashed line marks the fixed factor 1.25; the dotted line
marks the pooled factor 1.118 selected across all corpora and
size combinations.
The factors shown here are selected using the observations in
their respective combinations. Table~\ref{tab:confirmation}
reports their pooled and cross-corpus calibration alternatives
on disjoint source collections.}
\label{fig:multiplier-transfer}
\end{figure}

These comparisons were designed after both experiments had
been scored. Each factor is selected using only the calibration
outcomes specified below.
For forecast-to-observation ratios $r_j=F_j/\bar D_j$, a
constant minimizing mean absolute log error is
\[
 a=\exp\left\{-\operatorname{median}_j\log r_j\right\}.
\]
Banks, size combinations, and corpora receive equal weight.
We evaluate a sweep from 0.85 to 1.50 in increments of 0.005.
Figure~\ref{fig:multiplier-transfer} shows the exact optimum
for each size combination, computed from the median formula.

Selecting one factor from the full twenty-pair grid gives
$a=1.118$. Applied unchanged to the six-pair grid, this factor
gives error 0.045, compared with 0.060 for the jackknife.
The paired difference is
$-0.014\ [-0.020,-0.009]$.
Selecting the factor using only the twenty-pair $m<n$ subset
gives $a=1.222$ and error 0.030 at $m=1024,n=2048$ in the
six-pair experiment.

To test transfer across corpora, we select a factor using two
corpora from the twenty-pair experiment and evaluate it on the
third corpus in the six-pair experiment.
The factors for TinyStories, WikiText-103, and CNN/DailyMail
are 1.086, 1.116, and 1.130, respectively.
Their pooled error on the six-pair grid is 0.057, with paired
difference from the jackknife
$-0.003\ [-0.008,0.002]$.

Two further comparisons use the twenty-pair grid to test
transfer across sizes. We exclude either one target size or
one ratio $m/n$ when selecting the factor, then evaluate it
on the excluded combinations.
Both procedures give error 0.062, compared with 0.061 for
the jackknife. The paired differences are
$0.001\ [-0.003,0.006]$ and
$0.001\ [-0.004,0.005]$, respectively.

All intervals use 5,000 resamples of whole reference banks
and A/B pairs. Measurements at nested sizes are carried
together, and the factors are selected again within each
resample. The twenty-pair and six-pair experiments are
resampled independently.
The size-based comparisons separate the calibration and
evaluation combinations, but use shared banks and A/B pairs.

So a fitted multiplier does about as well as the jackknife, but only when past A/B disagreements are available to fit it; the jackknife needs none.

\subsection{Data allocation and shared quantities}
\label{app:computation}
The output vocabulary consists of GPT-2's 50,257 token IDs,
including EOS; MASK and PAD are excluded. We use 64-token chunks
and fix $R=2$, $\alpha=5$, and auxiliary ridge $\rho=0.01$.
A source is a story in TinyStories, a paragraph in WikiText-103,
or an article in CNN/DailyMail. Each news article contributes at
most two spaced chunks. Because the cached WikiText token IDs
do not mark article boundaries, source grouping accounts for
dependence within paragraphs only.

The 2,048-chunk reference banks used in the seven-pair comparison
contain 768 stories, 895 paragraphs, and 1,025 articles, respectively.
This comparison includes mask rates $q\in\{0.2,0.5,0.8\}$.
The three-pair comparison at $n=4096$ and $8192$ uses a separate
auxiliary allocation, which also supplies the finite-reference
correction and neural comparisons. These comparisons share
evaluation text and saved masks.

Within each auxiliary allocation, one split estimates $p$ and
the conditional rows $\widehat P^{\rm aux}_\delta$. A separate
split determines the weights using the centered predictors
\[
 u_{\delta,i}
 =\widehat P^{\rm aux}_\delta(\cdot\mid X_{i-\delta})-p
\]
at mask rate $q=0.5$, with unavailable predictors set to zero.
We pool inner products between predictors into a Gram matrix $G$
and their inner products with $e_{X_i}-p$ into a vector $b$.
For a nonempty set $\cA$ of visible offsets, the weights are
\[
 w_{\cA}
 =
 \left[
 G_{\cA\cA}
 +\rho\max\left\{
 \frac{\tr(G_{\cA\cA})}{|\cA|},10^{-12}
 \right\}I
 \right]^{-1}b_{\cA}.
\]
When $\cA$ is empty, the reconstruction is $p$. Within each
experiment, the weights are fixed across corpus sizes and mask
rates. Radius, smoothing, and ridge are chosen using development
data, independently of the A/B disagreement used for evaluation.

Each real-text evaluation uses 128 chunks with one saved Bernoulli
mask per chunk at the specified mask rate. If a mask hides no
positions, we hide position zero. At $q=0.2$, the probability of
this event is $0.8^{64}<10^{-6}$. We first average disagreement
over hidden positions within each chunk, then average across
chunks with equal weight.

\paragraph{Multinomial implementation.}
\label{app:matched-disagreement}
For a reference bank of $m$ chunks and a target corpus size of
$n$ chunks, we substitute
\[
 \widetilde p_\delta(a\mid b)
 =
 \frac{C^{\rm ref}_\delta(b,a)+\alpha p_a}
 {C^{\rm ref}_\delta(b,+)+\alpha},
 \qquad
 \widetilde N_{\delta,n}(b)
 =
 \frac{n}{m}C^{\rm ref}_\delta(b,+)
\]
into \eqref{eq:multinomial-law}. The conditional estimate is fixed
as $n$ changes. We compute its collision probability,
$\sum_a\widetilde p_\delta(a\mid b)^2$, over the full vocabulary,
including tokens absent from the reference row.

The same-lag source forecast instead sums squared influence
contributions separately for each lag. This removes cross-lag
covariance terms while accounting for within-source dependence
at each lag, random row totals, and variation in source length.

\subsection{Disagreement across corpus and reference sizes}
\label{app:earlier-studies}
\label{app:reference-ratio}
\label{app:fresh-replication}
\label{app:extended-size}
\label{app:matched-disagreement-large}
\label{app:matched-reference-size}
\label{app:matched-size-response}
In the seven-pair comparison, observed disagreement decreases by
5--40\% across corpora as $n$ increases from 512 to 2,048 chunks.
A $1/n$ law would predict a 75\% decrease over this range.
The forecast based on a 2,048-chunk reference underpredicts
disagreement by 13--18\% at $n=2048$. At $n=4096$, the shortfall
increases to 27--33\%; this comparison uses seven pairs for
TinyStories, five for WikiText-103, and four for CNN/DailyMail.
Increasing the reference to 6,144 chunks reduces the shortfall
at $n=4096$ to 8--11\%, but leads to overprediction at $n=512$.

The comparison at $n=4096$ and $8192$ uses three A/B pairs per
corpus to examine both calibration and the decrease in
disagreement with corpus size. Holding the A/B outcomes fixed,
the multinomial forecast is closer to observed disagreement
in all six corpus--size conditions at each of $m=2048$ and
$4096$. At $m=8192$, the full source forecast is closer in four
of the six conditions, including all three corpora at $n=4096$.
The multinomial forecast nevertheless describes the percentage
decrease between the two corpus sizes more accurately.
Reference size can therefore change which method is better
calibrated without producing the same ranking for size response.

A reference-size comparison uses three independently sampled
A/B pairs at $n=4096$. Here we report calibration as the forecast
divided by the arithmetic mean of observed disagreement.
Increasing $m$ from 4,096 to 12,288 changes the source forecast
ratios from $(0.845,0.838,0.861)$ to $(1.010,1.012,0.988)$,
in the order TinyStories, WikiText-103, and CNN/DailyMail.
A separate 4,096-chunk reference bank gives ratios
$(0.846,0.859,0.882)$. Thus, the ratio
$m/n$ alone does not determine calibration.

This sensitivity to reference estimation motivates the
finite-reference correction experiments with multiple independent
banks.

\FloatBarrier
\section{Neural Protocol and Transfer Test}
\label{app:mdlm-results}

\subsection{Training and evaluation}
\begin{figure}[htbp]
\centering
\includegraphics[width=\textwidth]{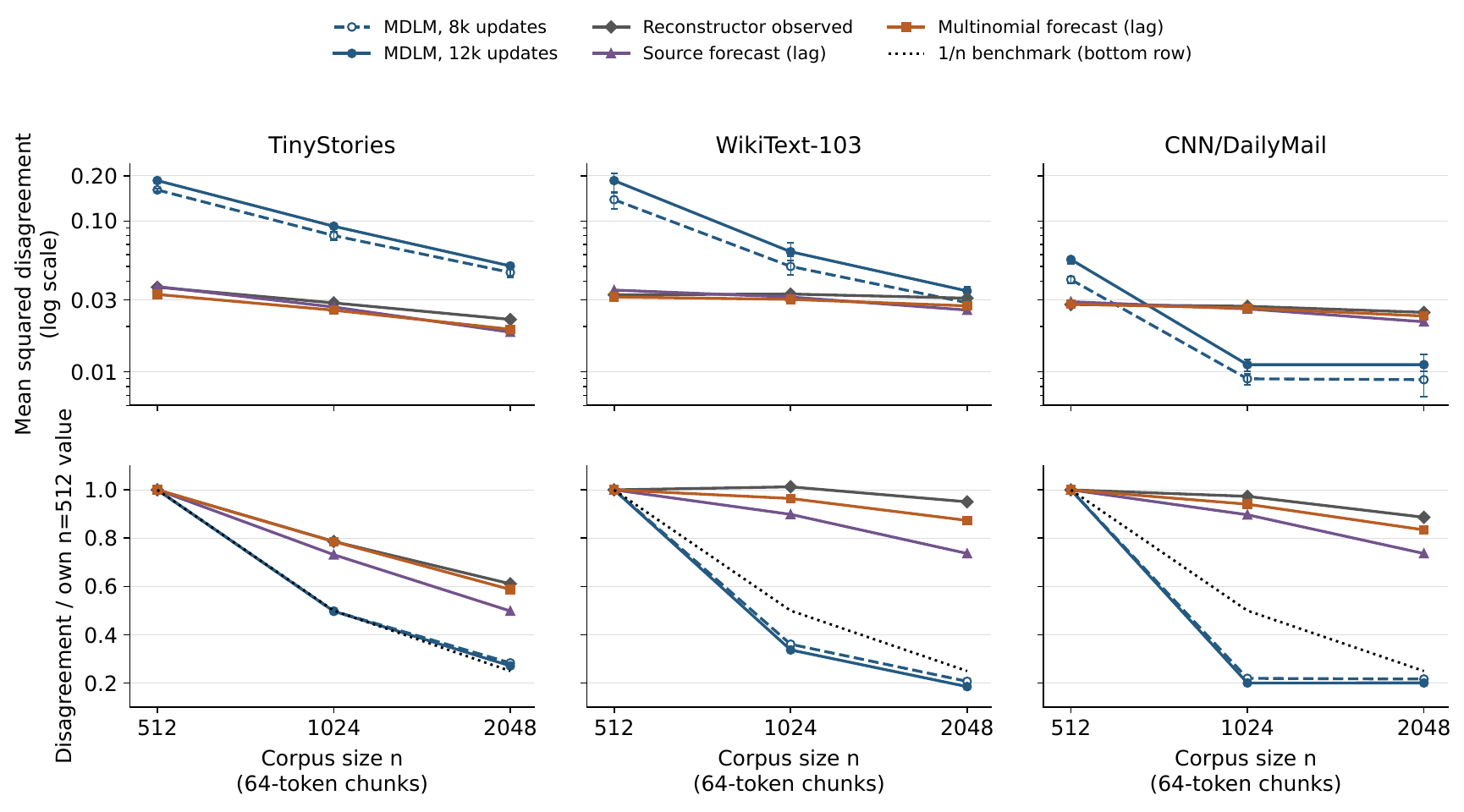}
\caption{Corpus-size response on the matched task at $q=0.5$.
The upper row shows observed mean disagreement for MDLM and the
reconstructor, together with reference forecasts for the reconstructor.
Both forecasts use 2,048 reference chunks. The vertical axis is
logarithmic, and bars span the three observed pair means. The lower
row divides each curve by its own $n=512$ value, with a dotted
$512/n$ benchmark starting at that value. Sizes count 64-token
chunks, and MDLM curves include both training stages.}
\label{fig:mdlm-disagreement}
\end{figure}

For each corpus, we train on three A/B pairs at each size
$n\in\{512,1024,2048\}$. The neural and count-based comparisons
use the same vocabulary, 128 evaluation chunks, and saved
$q=0.5$ masks. Count forecasts are computed from a 2,048-chunk
reference bank. Figure~\ref{fig:mdlm-disagreement} reports neural
disagreement after 8,000 and 12,000 updates.

The neural model is a bidirectional Transformer with two layers,
hidden width 96, four attention heads, feed-forward width 384,
tied input and output embeddings, and no dropout. It has
5,179,539 parameters. We train it using the masked-token objective
of MDLM \citep{sahoo2024mdlm}, within the discrete-diffusion
framework \citep{austin2021structured,lou2024ratios}.
We sample $t\sim\operatorname{Uniform}[0.05,0.95]$ and independently
hide each token with probability $t$. The time-conditioned model
minimizes
\begin{equation}
 \E_{X,t,M}\left[
 -\frac{1}{64t}
 \sum_{i\in H}\log f_{\theta,i}(X_i\mid Z,t)
 \right].
\label{eq:mdlm-training}
\end{equation}
Here, $f_{\theta,i}(X_i\mid Z,t)$ is the probability that the
Transformer with parameters $\theta$ assigns to the true token
$X_i$ at position $i$, given the masked input $Z$ and diffusion
time $t$. Training uses the MDLM loss over the specified interval of
masking probabilities. The output softmax covers the 50,257-token
vocabulary, excluding MASK and PAD. A training example with no
hidden positions contributes zero loss.

For each corpus, a model trained on auxiliary data for 4,000
updates provides the common initialization for the A/B runs.
Both models begin with newly initialized AdamW optimizer states.
Within each pair, they use identical minibatch indices, diffusion
times, training masks, learning-rate schedules, and update counts.
Each optimizer update accumulates gradients over four
microbatches of four chunks, giving an effective batch size of 16.
The learning rate increases to $5\times10^{-4}$ over the first
400 updates, decreases to $5\times10^{-5}$ at update 8,000, and
remains at that value through update 12,000. AdamW uses
$(\beta_1,\beta_2)=(0.9,0.98)$, weight decay $0.01$, and gradient
clipping at one.

Evaluation uses the saved masks at diffusion time $t=q=0.5$.
For a model class $h$, we estimate disagreement by
\begin{equation}
 \widehat{\Dbar}^{\rm obs,h}_{n,q}
 =
 \frac{1}{3\cdot128}
 \sum_{r=1}^{3}\sum_{e=1}^{128}
 \|h_{A_r}(Z_e)-h_{B_r}(Z_e)\|_{M_e}^{2}.
\label{eq:mdlm-observed}
\end{equation}
This can also be written as
\[
 \widehat{\Dbar}^{\rm obs,h}_{n,q}
 =
 \frac{1}{3\cdot128}
 \sum_{r=1}^{3}\sum_{e=1}^{128}
 \frac{1}{|H_e|}
 \sum_{i\in H_e}
 \|h_{A_r,i}(Z_e)-h_{B_r,i}(Z_e)\|_2^2.
\]
Here, $M_e$ denotes the mask applied to evaluation chunk $e$,
and $H_e$ is its set of hidden positions. The norm
$\|\cdot\|_{M_e}^2$ averages squared Euclidean output differences
over positions in $H_e$.
This comparison measures variation across corpora conditional on
the shared initialization, training randomness, and update budget.
Under independent A/B corpus sampling, the corresponding
population disagreement satisfies the equal-size identity
\eqref{eq:identity}.

The two predictors have different information and output
constraints: MDLM uses the full visible chunk and returns token
probabilities, whereas the reconstructor uses the four offsets
$\{-2,-1,1,2\}$ and returns affine scores.

To assess sensitivity to training randomness, we also retrain
the A corpus of pair 1 with an alternative training seed at every
size. At $n=2048$ and 12,000 updates, disagreement between models
trained on this same corpus is 0.0281 for TinyStories, 0.0138 for
WikiText-103, and 0.0052 for CNN/DailyMail. These values measure
a different source of variation from the A/B comparison and are
reported separately, without subtraction from A/B disagreement.

For input-ranking comparisons, we compute Spearman correlations
over the 128 evaluation chunks within each pair and report the
mean of the three coefficients. We report them without significance tests, because chunks from the same source are not independent.

\subsection{Directional overlap and the frequency control}
\label{app:mdlm-subspaces}
Scalar disagreement measures the total size of an output
difference. We also ask whether neural and count-based differences
concentrate in similar directions in vocabulary space.

Let $Q\in\mathbb R^{50257\times64}$ have fixed, randomly chosen
orthonormal columns satisfying $Q^\top\mathbf1=0$.
For a given A/B pair and model class $h$, write
$\Delta_i^h(Z)=h_{A,i}(Z)-h_{B,i}(Z)$.
On a set $T$ of evaluation chunks, define
\[
 C_{h,T}
 =
 \frac{1}{|T|}
 \sum_{j\in T}\frac{1}{|H_j|}
 \sum_{i\in H_j}
 \bigl(Q^\top\Delta_i^h(Z_j)\bigr)
 \bigl(Q^\top\Delta_i^h(Z_j)\bigr)^\top.
\]
This is an uncentered second-moment matrix. Its trace is the
projected disagreement, including the contribution from the
mean output difference.

We partition the 128 evaluation chunks into two groups,
$T_1$ and $T_2$, of 64 chunks each, placing all chunks from a
source in the same group. For each group $T_f$, we use the other
group to select the four leading eigenvectors of the
reconstructor's second-moment matrix. Denote these orthonormal
directions by the columns of $U^{(-f)}$. The fraction of neural
disagreement in these directions, measured on the held-out
groups, is
\[
 F
 =
 \frac{
 \sum_{f=1}^2 |T_f|
 \tr\!\left\{
 (U^{(-f)})^\top C_{\mathrm{MDLM},T_f}U^{(-f)}
 \right\}
 }{
 \sum_{f=1}^2 |T_f|\tr(C_{\mathrm{MDLM},T_f})
 }.
\]
We compute this fraction for each of the three pairs at
$n=2048$ and 12,000 updates, then average across pairs.
The projection $Q$, the rank of four, and the source partition
were fixed after the scalar analyses and before computing
directional overlap. All reported fractions refer to energy
within the 64-dimensional projection.

A uniformly random four-dimensional subspace captures
$4/64=6.25\%$ of this energy in expectation. To determine whether
the reconstructor contributes more than a preference for frequent
tokens, we also use the four leading eigenvectors of
\[
 C_{\rm freq}
 =
 Q^\top\{\operatorname{diag}(p)-pp^\top\}Q.
\]
This matrix is the projected covariance of a one-hot token drawn
from the auxiliary unigram distribution. Its leading directions
capture at least as much neural disagreement as those selected
by the reconstructor (Section~\ref{sec:mdlm-subspaces}).
The control therefore measures how much directional overlap can
be explained by token frequencies alone. It does not compare
two unigram predictors: both would return the same $p$ and have
zero disagreement.

\subsection{Relation to the image diffusion setting}
\label{app:image-text-comparison}
The common question in the image and text settings is how
predictions change when the training data are sampled again.
In the image analysis of \citet{wang2026rmt}, the tractable
predictor is the empirical linear denoiser
\[
 \widehat D_\sigma(x)
 =
 \widehat\mu
 +
 \widehat\Sigma
 (\widehat\Sigma+\sigma^2I)^{-1}
 (x-\widehat\mu).
\]
Its dependence on the dataset is determined by the empirical
mean and covariance. The text reconstructor instead depends on
lag counts through smoothed conditional rows. For fixed tables
and masks, it is affine in the visible one-hot tokens, while
row normalization makes it nonlinear in the sampled counts.
Related local co-occurrence statistics appear in word association
\citep{church1990word}, GloVe \citep{pennington2014glove}, and
the matrix interpretation of skip-gram with negative sampling
\citep{levy2014implicit}.
Table~\ref{tab:image-text-map} summarizes the correspondence.
The image analysis includes both denoising at a fixed noise
level and complete sampling maps. Our text comparisons evaluate
a single masked-input reconstruction, using either affine scores
or neural token probabilities.

\begin{table}[htbp]
\centering
\caption{Comparison of the image diffusion setting of
\citet{wang2026rmt} with the two text predictors. Each setting
measures variation across datasets while holding the evaluation
input fixed. The image theory analyzes a linear denoiser and
its sampling map; the text theory analyzes normalized lag
counts and their reconstruction scores. MDLM provides a neural
test of the count-based forecasts.}
\label{tab:image-text-map}
\small
\setlength{\tabcolsep}{4pt}
\renewcommand{\arraystretch}{1.12}
\begin{tabular}{@{}
 >{\raggedright\arraybackslash}p{0.12\textwidth}
 >{\raggedright\arraybackslash}p{0.28\textwidth}
 >{\raggedright\arraybackslash}p{0.27\textwidth}
 >{\raggedright\arraybackslash}p{0.27\textwidth}
 @{}}
\toprule
Quantity
& Image diffusion
& Text reconstructor
& Trained MDLM \\
\midrule
Data sampling
& Images sampled from a distribution
& Sources supplying token chunks
& The same source-disjoint A/B corpora \\
Corruption
& Additive Gaussian noise with variance $\sigma^2$
& Independent token masking with probability $q$
& Absorbing token masks; evaluation at $t=q=0.5$ \\
Dependence on data
& Empirical mean $\widehat\mu$ and covariance
  $\widehat\Sigma$ for the linear denoiser
& Lag counts and conditional rows; auxiliary $p$ and weights fixed
& Transformer parameters learned from masked chunks \\
Predictor
& Empirical linear denoiser $\widehat D_\sigma(x)$
& Weighted conditional rows in \eqref{eq:rule}
& Time-conditioned Transformer probabilities \\
Population target
& Output variation over independently sampled datasets
& $\Dbar^{\rm lag}_{n,q}$ over source collections
& $\Dbar^{\rm MDLM}_{n,q}$ conditional on shared training randomness \\
Shared input
& Noisy image for denoising; initial noise for sampling
& Identical text and mask
& Identical text, mask, and diffusion time \\
Analysis
& Random-matrix expressions for fluctuations of the linear
  denoiser and sampling map
& Source-count covariance or multinomial row variance
& Observed A/B disagreement compared with count-based forecasts \\
Outputs compared
& Denoiser outputs and generated samples
& One-step reconstruction scores
& One-step token probabilities \\
\bottomrule
\end{tabular}
\end{table}

\FloatBarrier
\section{\texorpdfstring{Reconstruction Quality and Top-$k$ Accuracy}{Reconstruction Quality and Top-k Accuracy}}
\label{app:quality}

\subsection{Relative MSE improvement}
Across the 54-condition grid, the reconstructor's mean relative
MSE improvement over the auxiliary unigram is 13.16\% for
TinyStories, 5.19\% for WikiText-103, and 3.95\% for CNN/DailyMail.

We examine the role of local order with a separate shuffle
control at $n=512,q=0.5$. Shuffling tokens within each chunk
preserves token counts and source membership, but yields
relative MSE improvements of only 0.0042\%, 0.0003\%, and
0.0001\%, respectively. These near-zero gains indicate that
the reconstructor depends strongly on local token order.
The control evaluates a single size and mask rate, whereas
the percentages above average over the full grid.

Table~\ref{tab:mdlm-quality} compares the reconstructor and MDLM
on the same evaluation inputs at $n=2048,q=0.5$. At both neural
checkpoints, MDLM achieves lower MSE and higher top-1 and top-64
accuracy on TinyStories. The reconstructor performs better on
all three measures for WikiText-103 and CNN/DailyMail.
At $n=512$, MDLM's relative MSE improvement is negative on
WikiText-103 and CNN/DailyMail after both 8,000 and 12,000 updates.
For CNN/DailyMail, it is also slightly negative at $n=1024$.

\begin{table}[!htbp]
\centering
\caption{Reconstruction quality at $n=2048,q=0.5$ on the shared
128 evaluation chunks. Each row averages six models from three
A/B pairs. MSE sums squared error over the 50,257 vocabulary
coordinates, averages over hidden positions within each chunk,
and then weights chunks equally. MSE gain is the percentage
reduction relative to the auxiliary unigram. Top-1 and top-64
accuracy are also reported as percentages. Arrows indicate the
preferred direction.}
\label{tab:mdlm-quality}
\small
\setlength{\tabcolsep}{6pt}
\begin{tabular}{@{}llrrrr@{}}
\toprule
Corpus & Method
& MSE $\downarrow$
& MSE gain (\%) $\uparrow$
& Top-1 (\%) $\uparrow$
& Top-64 (\%) $\uparrow$ \\
\midrule
TinyStories & Reconstructor & 0.8450 & 14.34 & 28.12 & 75.80 \\
TinyStories & MDLM, 8k & 0.7855 & 20.37 & 33.14 & 78.75 \\
TinyStories & MDLM, 12k & 0.7811 & 20.82 & 33.88 & 79.15 \\
\addlinespace
WikiText-103 & Reconstructor & 0.9372 & 5.45 & 15.75 & 48.49 \\
WikiText-103 & MDLM, 8k & 0.9603 & 3.12 & 12.09 & 44.66 \\
WikiText-103 & MDLM, 12k & 0.9599 & 3.16 & 12.37 & 44.39 \\
\addlinespace
CNN/DailyMail & Reconstructor & 0.9482 & 4.43 & 13.94 & 52.11 \\
CNN/DailyMail & MDLM, 8k & 0.9821 & 1.01 & 7.38 & 45.45 \\
CNN/DailyMail & MDLM, 12k & 0.9801 & 1.21 & 7.86 & 45.40 \\
\bottomrule
\end{tabular}
\end{table}

\subsection{\texorpdfstring{Top-$k$ accuracy}{Top-k accuracy}}
\label{app:topk-reconstructor}
Squared error evaluates the full output vector. Top-$k$ accuracy
instead asks whether the true token appears among the $k$
highest-scoring vocabulary entries. We rank all 50,257 tokens,
including EOS, using the reconstructor's raw affine scores or
the neural model's logits. Ties are resolved by ascending token
ID. As with MSE, we average over hidden positions within each
chunk and then give chunks equal weight.

To compare these rankings with the same auxiliary unigram used
for MSE, write $\operatorname{Acc}_k$ as a fraction and define
\[
 \mathcal E_k(f)=1-\operatorname{Acc}_k(f),\qquad
 \operatorname{RelErrRed}_k(f)
 =1-\frac{\mathcal E_k(f)}{\mathcal E_k(p)}.
\]
The relative error reduction measures the fraction of the
unigram's top-$k$ errors removed by $f$.

\begin{figure}[htbp]
\centering
\includegraphics[width=\textwidth]{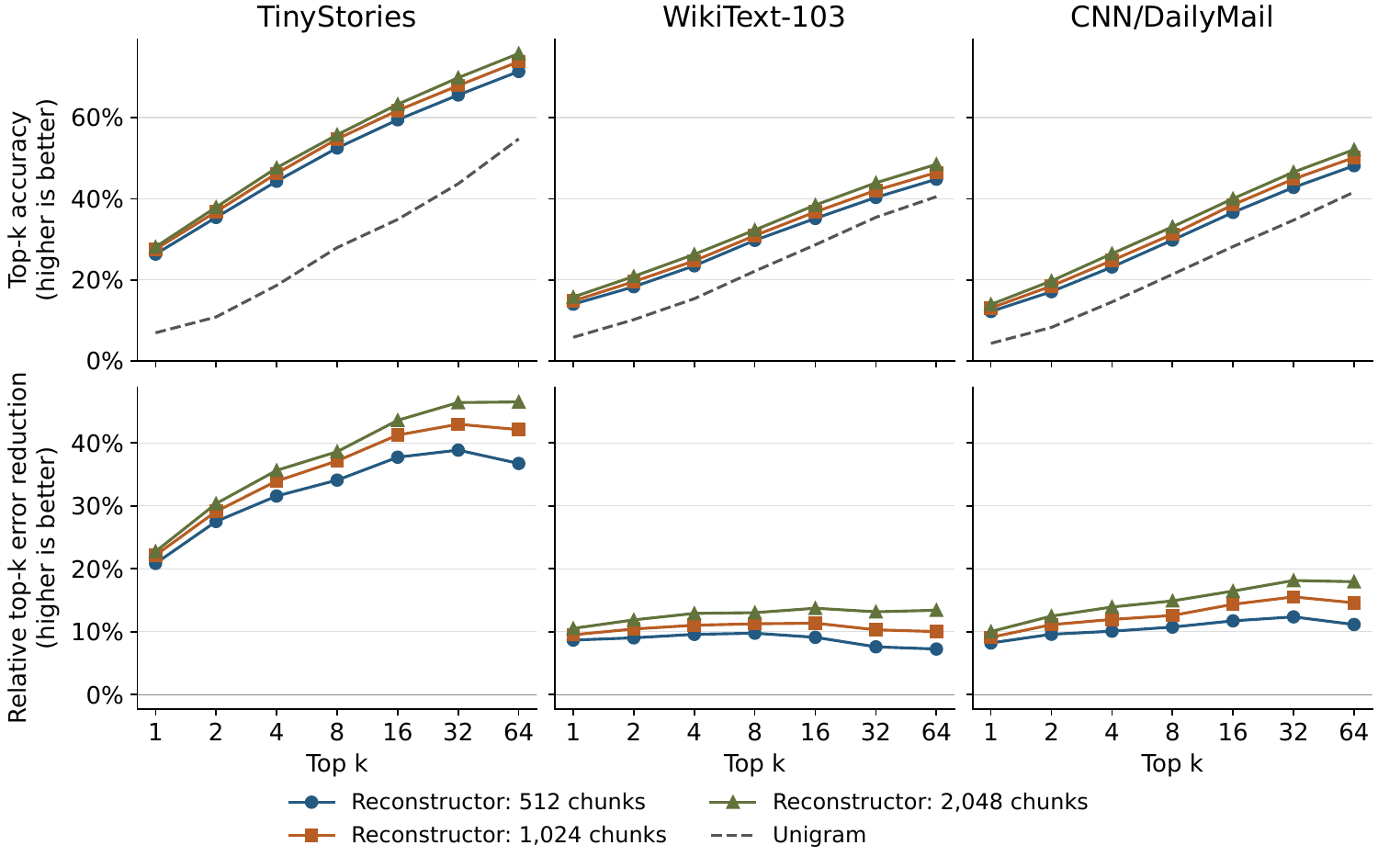}
\caption{Top-$k$ reconstruction quality at $q=0.5$.
The upper row compares the radius-two reconstructor at three
corpus sizes with the auxiliary unigram. The lower row gives
the relative reduction in top-$k$ error against that same
unigram. Colors and markers denote corpus size in both rows.
Each reconstructor curve averages A and B within each pair,
then averages the three pairs. Higher values indicate better
performance in both rows.}
\label{fig:lag-topk}
\end{figure}

Figure~\ref{fig:lag-topk} reports the reconstructor's accuracy
and relative error reduction at seven cutoffs and three corpus
sizes. At $q=0.5$, its accuracy exceeds the unigram's at every
tested size and cutoff. Although accuracy is nondecreasing in
$k$, relative error reduction need not be, because both the
reconstructor's and the unigram's error rates change with the
cutoff. For CNN/DailyMail, relative error reduction peaks at
$k=32$ for all three sizes. For WikiText-103, the values at
$k=32$ and $64$ are below the value at $k=16$.

\FloatBarrier
\section{Calibration Checks and Smoothing}
\label{app:diagnostics}

\subsection{Controlled population and error decomposition}
\label{app:missing-contexts}

We use a population with known moments to separate reference
estimation error from linearization error. The control fixes
$K=512$, $L=8$, $R=2$, $q=0.5$, and ridge $0.01$.
Each of 32 topics selects 32 distinct tokens using weights
proportional to $k^{-1.05}$. Its reset distribution $v_h$ assigns
weight $0.55$ to the uniform distribution on these tokens and
$0.45$ to the global Zipf distribution. Each source chooses a
topic uniformly. Chunks begin with a draw from $v_h$ and then
follow
\[
 P_h=rS+(1-r)\mathbf1v_h^\top,
 \qquad r=0.25,
\]
where $S$ maps each token to its cyclic successor. We compare
independent tokens, dependent tokens in one-chunk sources,
and four chunks sharing a topic within each source. The
independent-token control sets both the topic weight and $r$
to zero.

For each control, we evaluate 80 A/B pairs at each
$n\in\{256,1024\}$ and four reference banks at each
$m\in\{128,512,2048\}$. Independent auxiliary samples provide
4,096 chunks for estimating $p$ and two sets of 1,024 chunks
for determining weights. Evaluation uses 32 separate chunks
with fixed masks; empty masks contribute zero. Smoothing
defaults to $\alpha=5$. For the four-chunk sources, we also
test $\alpha=1,20$ using the same source draws, auxiliary data,
and masks, with weights recomputed for each smoothing value.

Let $D_{\rm obs}$ denote mean A/B disagreement for the nonlinear
reconstructor and $D_{\rm lin}$ its counterpart after linearizing
at the population mean counts. For a fixed reference bank,
write the linearized output difference as $U+V$, where $U$
contains contributions from rows with zero reference totals
and $V$ contains those from observed rows. Define
\[
 D_U=\E\|U\|_M^2,\qquad
 D_V=\E\|V\|_M^2,\qquad
 C_{UV}=\E\langle U,V\rangle_M.
\]
Since $D_{\rm lin}=D_U+D_V+2C_{UV}$, the error of a reference
forecast $F_{\rm ref}$ decomposes exactly as
\begin{equation}
 F_{\rm ref}-D_{\rm obs}
 =
 \underbrace{-(D_U+2C_{UV})}
 _{\text{unseen rows and cross terms}}
 +
 \underbrace{(F_{\rm ref}-D_V)}
 _{\text{reference error on observed rows}}
 +
 \underbrace{(D_{\rm lin}-D_{\rm obs})}
 _{\text{linearization error}}.
\label{eq:missing-decomposition}
\end{equation}
The cross term can have either sign, so the contribution of
unseen rows cannot be described by $D_U$ alone.

For four chunks per source at $\alpha=5$, $n=1024$, and $m=128$,
the three terms divided by $D_{\rm obs}$ are $-32.6\%$,
$-38.6\%$, and $+3.2\%$, respectively. Missing rows therefore
explain only part of the underprediction. Reference error on
observed rows contributes substantially, while linearization
error partly offsets the reference error.

\subsection{Population forecasts and sensitivity to smoothing}
\label{app:population-calibration}
\label{app:population-oracle}

The generator's exact mean lag-count matrix per chunk is
\begin{equation}
 m_\delta
 =
 \frac{1}{32}\sum_{h=1}^{32}
 \sum_{t=0}^{L-\delta-1}
 \operatorname{diag}(v_h^\top P_h^t)P_h^\delta,
 \qquad
 m_{-\delta}=m_\delta^\top,\quad \delta>0.
\label{eq:population-counts}
\end{equation}
For the source forecast, we evaluate the derivative at
$B_{\delta,n}=nm_\delta$ and use the exact joint covariance
of the counts at size $n$. This covariance includes overlapping
pairs, cross-lag correlations, and dependence between chunks
that share a topic. Products of pair counts involve at most
four token positions, whose joint probabilities can be computed
exactly from the Markov chains and averaged over topics.
We check the moment calculation by enumeration in a population
with three tokens and length-four chunks.

The multinomial forecast uses the same population mean row
totals and the unsmoothed population conditional probabilities
in its collision term. Both forecasts use the same fixed
auxiliary quantities and evaluation inputs.

\begin{table}[!htbp]
\centering
\caption{Calibration using exact population moments.
Entries are forecasts divided by mean nonlinear A/B disagreement;
values closer to one indicate better agreement. Brackets give
95\% intervals from 5,000 bootstrap resamples of the 80 A/B pairs,
with forecasts, auxiliary quantities, and evaluation inputs
fixed. The first six rows use $\alpha=5$. The smoothing
comparisons use four chunks per source and the same source
draws across values of $\alpha$.}
\label{tab:population-oracle}
\small
\setlength{\tabcolsep}{8pt}
\begin{tabular}{@{}lrrr@{}}
\toprule
Control & $n$ & Source & Multinomial \\
\midrule
Independent tokens & 256 & 1.073 [1.062, 1.085] & 1.069 [1.058, 1.081] \\
Independent tokens & 1,024 & 1.035 [1.024, 1.045] & 1.033 [1.022, 1.043] \\
Dependent tokens & 256 & 1.092 [1.076, 1.108] & 1.051 [1.036, 1.067] \\
Dependent tokens & 1,024 & 1.046 [1.032, 1.060] & 1.010 [0.997, 1.023] \\
Four chunks, $\alpha=5$ & 256 & 1.117 [1.099, 1.135] & 1.065 [1.048, 1.082] \\
Four chunks, $\alpha=5$ & 1,024 & 1.027 [1.015, 1.040] & 0.991 [0.979, 1.003] \\
\addlinespace
Four chunks, $\alpha=1$ & 256 & 1.201 [1.176, 1.225] & 1.162 [1.139, 1.185] \\
Four chunks, $\alpha=1$ & 1,024 & 0.977 [0.960, 0.995] & 0.962 [0.945, 0.979] \\
Four chunks, $\alpha=20$ & 256 & 1.047 [1.029, 1.065] & 0.995 [0.978, 1.012] \\
Four chunks, $\alpha=20$ & 1,024 & 1.025 [1.013, 1.037] & 0.976 [0.965, 0.988] \\
\bottomrule
\end{tabular}
\end{table}

In all six $\alpha=5$ settings, the multinomial forecast is
closer to observed disagreement even after reference estimation
error has been removed (Table~\ref{tab:population-oracle}).
The source forecast overpredicts in these settings, and the
smaller multinomial forecast partly offsets this positive
linearization error. For four chunks per source at
$\alpha=1,n=1024$, however, the source forecast already
underpredicts, so the smaller multinomial forecast increases
the shortfall.

So in these simulations the multinomial law looks better partly because errors cancel. This does not show that the same happens on real text.

\subsection{Smoothing and rare contexts on real text}
\label{app:occupancy}
\label{app:alpha-summary}

Table~\ref{tab:main-alpha-ablation} examines smoothing using the
same 20 A/B pairs and six reference banks per corpus as
Appendix~\ref{app:correction-study}. We use a vocabulary of
50,257 tokens, 64-token chunks, $R=2$, and $q=0.5$.
The 128 evaluation chunks, masks, and auxiliary unigram are
fixed across smoothing values. Reconstruction weights are
also fixed at their $\alpha=5$ values, so the comparison isolates
changes in the conditional rows. The full sweep uses
$\alpha\in\{0.5,1,2,5,10,20,50\}$; the table reports
$\alpha=1,5,20$. The data allocations and $\alpha=5$ outcomes
were already known. Forecasts for the other smoothing values
were saved before scoring their corresponding A/B outcomes.

Both the source and multinomial forecasts use the source-split
jackknife in \eqref{eq:reference-jackknife}, with each component
recomputed at the chosen $\alpha$ and target size. For each
corpus, we average $|\log(F_{b,m,n}/\bar D_n)|$ equally over
six banks and the three conditions
\[
 (m,n)\in\{(1024,2048),(1024,4096),(2048,4096)\}.
\]
Here, $\bar D_n$ is the mean disagreement across the 20 pairs
at the chosen $\alpha$. Lower error indicates better calibration.
The numerical records include paired intervals from 5,000
bootstrap resamples of whole banks and A/B pairs. Each
resampled bank or pair carries its measurements across methods,
smoothing values, and nested sizes. These intervals condition
on the auxiliary data and evaluation inputs.

After jackknife correction, the multinomial forecast has lower
error at $\alpha=1,5$ in all three corpora. At $\alpha=20$,
the source forecast is more accurate on TinyStories and
WikiText-103, while both CNN/DailyMail errors round to $0.024$.
So the multinomial law's advantage depends on $\alpha$. The simulations show that errors can cancel, but they do not explain why the ranking flips at $\alpha=20$ on real text.

A row with context count $N$ assigns weight $\alpha/(N+\alpha)$
to the shared unigram. Smoothing is therefore substantial when
$N$ is small relative to $\alpha$; an unseen row returns $p$
for every $\alpha>0$. In one 2,048-chunk WikiText-103 reference
bank, 7.3\% of evaluation queries address unseen rows, and
20.8\% address rows with fewer than five observations.
These frequencies count repeated queries and describe that
single bank. Rare and unseen contexts connect this setting
to species-frequency estimation \citep{good1953population}
and statistical inference for rare text events
\citep{dunning1993accurate}.

Across $n=512,1024,2048,4096$, increasing $\alpha$ from 1 to 20
produces a shallower decrease in observed disagreement in all
three corpora. The paired 95\% intervals for the change in
log-log slope lie entirely above zero. Both the calibration
pattern and the flatter size response persist when weights
are re-estimated on the same auxiliary data for each $\alpha$.
The numerical records include every smoothing value, reference
size, and reconstruction-quality measurement.

\end{document}